\documentclass[lettersize,journal]{IEEEtran}
\usepackage{soul}
\DeclareRobustCommand{\hlcyan}[1]{{\sethlcolor{cyan}\hl{#1}}}
\soulregister\cite7
\soulregister\ref7
\renewcommand{\hlcyan}[1]{#1}

\usepackage{cite}

\ifCLASSINFOpdf
  \usepackage[pdftex]{graphicx}
\else
\fi
\usepackage{amsmath}
\usepackage{algorithm}[H]
\usepackage{algpseudocode}
\ifCLASSOPTIONcompsoc
 \usepackage[caption=false,font=normalsize,labelfont=sf,textfont=sf]{subfig}
\else
 \usepackage[caption=false,font=footnotesize]{subfig}
\fi
\usepackage{url}

\usepackage{amsfonts}
\usepackage{xcolor}
\usepackage{colortbl}
\usepackage{threeparttable}
\usepackage{multirow}
\usepackage{booktabs}
\usepackage[hidelinks]{hyperref}

\newcommand{\boldres}[1]{{\textbf{\textcolor{black}{#1}}}}
\newcommand{\secondres}[1]{{\underline{\textcolor{black}{#1}}}}
\newcommand\Tstrut{\rule{0pt}{2.6ex}}         
\newcommand\Bstrut{\rule[-0.9ex]{0pt}{0pt}}   

\begin{document}
%
\title{SageFormer: Series-Aware Framework for Long-Term Multivariate Time Series Forecasting}
%
%
%

\author{Zhenwei~Zhang,~\IEEEmembership{Student~Member,~IEEE,}
        Linghang~Meng,~\IEEEmembership{Student~Member,~IEEE,}
        Yuantao~Gu,~\IEEEmembership{Senior~Member,~IEEE}
\thanks{Manuscrpit submitted October 23, 2023; revised January 15, 2024. This work was supported in part by the National Natural Science Foundation of China under Grant U2230201, Grant from the Guoqiang Institute, Tsinghua University, and in part by the Clinical Medicine Development Fund of Tsinghua University. (Corresponding author: Yuantao Gu.)

The authors are with the Department of Electronic Engineering, Tsinghua University, Beijing 100084, China (e-mail: 
zzw20@mails.tsinghua.edu.cn; 
menglh95@163.com; 
gyt@tsinghua.edu.cn).}
}

%
%

\markboth{IEEE INTERNET OF THINGS JOURNAL}%
{Zhang \MakeLowercase{\textit{et al.}}: SageFormer: SERIES-AWARE FRAMEWORK}
%



\maketitle

\begin{abstract}
In the burgeoning ecosystem of Internet of Things, multivariate time series (MTS) data has become ubiquitous, highlighting the fundamental role of time series forecasting across numerous applications. The crucial challenge of long-term MTS forecasting requires adept models capable of capturing both intra- and inter-series dependencies. Recent advancements in deep learning, notably Transformers, have shown promise. However, many prevailing methods either marginalize inter-series dependencies or overlook them entirely. To bridge this gap, this paper introduces a novel series-aware framework, explicitly designed to emphasize the significance of such dependencies. At the heart of this framework lies our specific implementation: the SageFormer. As a Series-aware Graph-enhanced Transformer model, SageFormer proficiently discerns and models the intricate relationships between series using graph structures. Beyond capturing diverse temporal patterns, it also curtails redundant information across series. Notably, the series-aware framework seamlessly integrates with existing Transformer-based models, enriching their ability to comprehend inter-series relationships. Extensive experiments on real-world and synthetic datasets validate the superior performance of SageFormer against contemporary state-of-the-art approaches.
\end{abstract}


%
\IEEEpeerreviewmaketitle

\section{Introduction}
%
%
%
%

\IEEEPARstart{W}{ith the} rise of the Internet of Things (IoT), an ever-increasing number of interconnected devices have found their way into our daily lives, from smart homes and industries to healthcare and urban planning~\cite{zanella2014internet}. These devices continuously generate, exchange, and process large amounts of data, creating an intricate network of communication. Among the various forms of data produced, Multivariate Time Series (MTS) data emerges as a particularly prevalent and crucial type. Originating from the concurrent observations of multiple sensors or processors within IoT devices, MTS data paint a holistic picture of the complex interplays and temporal dynamics inherent to the IoT landscape~\cite{9802652, huang2022adaptive}.

Predicting the future behaviors within this burgeoning IoT-driven data environment is of paramount importance. Within IoT systems, forecasting MTS data optimizes operations and ensures security, especially in key sectors such as energy~\cite{9802652}, transportation~\cite{huang2022adaptive}, and weather~\cite{kashyap2021towards}.
While much of the current research accentuates the need for short-term forecasts to respond to immediate challenges, the realm of long-term forecasting stands as an equally significant frontier~\cite{haoyietal-informer-2021}. Long-term predictions offer insights into the larger temporal patterns and relationships within MTS data. However, they also bring forth their own set of complexities. Modeling over extended timeframes amplifies even minor errors, making the task considerably more challenging but undeniably valuable~\cite{haoyietal-informer-2021}. 

\begin{figure*}
    \centering
    \subfloat[Series-aware framework (Ours)]{
        \includegraphics[width=0.99\textwidth]{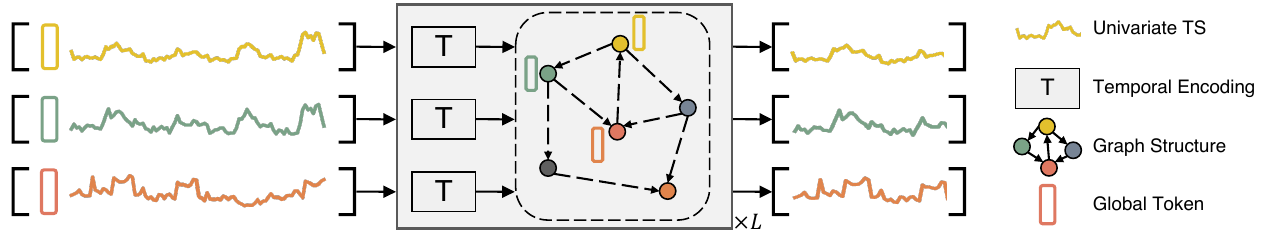}
        \label{fig:series-aware}
    }
    \\
    \subfloat[ Series-mixing framework]{
        \includegraphics[width=0.49\textwidth]{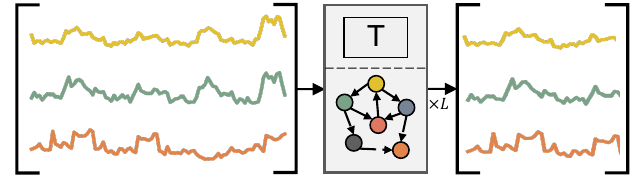}
        \label{fig:series-mixing}
    }
    \hfill
    \subfloat[ Series-independent framework]{
        \includegraphics[width=0.47\textwidth]{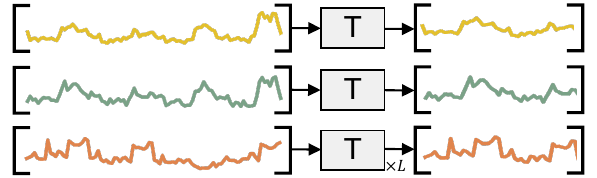}
        \label{fig:series-independent}
    }
    \caption{Illustration of three different ways of modeling series dependencies: (a) The proposed series-aware framework. Prior to the original input tokens into the Transformer encoder, we incorporate learnable global tokens to capture the intrinsic features of each series. The embedding tokens are processed through multiple SageFormer layers, where temporal encoding and graph aggregation are performed iteratively. (b) The series-mixing framework amalgamates all series at once and employs temporal encoding or GNNs to process them. (c) The series-independent framework handles each series separately, improving the learning of unique temporal patterns for different series.}
    \label{SageFormerCompare}
\end{figure*}

In recent years, deep learning methods~\cite{2018Modeling, lim2021time}, especially those employing Transformer architectures~\cite{2019Enhancing, haoyietal-informer-2021, wu2021autoformer, zhou2022fedformer}, have outperformed traditional techniques such as ARIMA and SSM~\cite{durbin2012time} in long-term MTS forecasting tasks. 
Many Transformer-based models predominantly focus on temporal dependencies and typically amalgamate varied series into hidden temporal embeddings using linear transformations, \hlcyan{termed the ``series-mixing framework''} (Fig.~\ref{fig:series-mixing}). However, within these temporal embeddings, inter-series dependencies are not explicitly modeled, leading to inefficiencies in information extraction~\cite{zhang2023crossformer}.
Interestingly, some recent studies~\cite{Zeng2022AreTE, nie2022time} have found that models that purposefully exclude inter-series dependencies, \hlcyan{termed the ``series-independent framework''} (Fig.~\ref{fig:series-independent}), can produce significantly improved prediction results due to their enhanced robustness against distribution drifts~\cite{han2023capacity}. However, this approach can be suboptimal for certain datasets for completely overlooking inter-series dependencies (see section~\ref{sec:cycleGraph}). This underscores the intricate balance needed in modeling both intra- and inter-series dependencies, marking it as a crucial area for MTS forecasting research. In this paper, we introduce \hlcyan{the ``series-aware framework''} (Fig.~\ref{fig:series-aware}) to bridge this research gap.


In this paper, we delve into the intricacies of inter-series dependencies in long-term MTS forecasting problems. We introduce the \textbf{series-aware framework} designed to precisely model inter-series dependencies, as illustrated by Fig.~\ref{fig:series-aware}. This framework forms the foundation upon which we develop the Series-Aware Graph-Enhanced Transformer (\textbf{SageFormer}), a series-aware Transformer model enhanced with graph neural networks (GNN). Learning with graph structures, we aim to distinguish series using interactable global tokens and improve the modeling ability for diverse temporal patterns across series through graph aggregation. The series-aware framework can function as a universal extension for Transformer-based structures, better utilizing both intra- and inter-series dependencies and achieving superior performance without greatly affecting model complexity. We contend that the proposed SageFormer addresses two challenges in long-term inter-series dependencies modeling:

\begin{enumerate}
    \item How can diverse temporal patterns among series be effectively represented?
\end{enumerate}

The series-aware framework augments the conventional series-independent framework. Our enhancement involves the introduction of several globally interactable tokens placed ahead of the input tokens. These global tokens are adept at capturing overarching information for each variable through intra-series interactions, employing mechanisms such as self-attention, CNNs, and the like. Moreover, they play a pivotal role in enabling information exchange between sequences via inter-series interactions. The global tokens enables series-aware framework to learn not only individual series' temporal patterns but also focus on dependencies between different series, thereby enhancing diversity and effectively addresses the limitations associated with the series-independent framework, as detailed in Section~\ref{sec:cycleGraph}.

\begin{enumerate}
    \setcounter{enumi}{1}
    \item How can the impact of redundant information across series be avoided?
\end{enumerate}


\hlcyan{In time series analysis, redundant information\cite{prichard1995generalized, doi:10.1142/S0218127496001879} primarily manifests in two forms: temporal and series redundancy. Temporal redundancy involves repetitive patterns within a single series, like periodicity or seasonality, that add no new information to forecasts. Series redundancy appears when multiple series in a dataset show similar patterns, typically due to correlation or co-movement between them. Acknowledging the detrimental impact of redundant information, as identified in existing literature\cite{shekar2018selection, li2023mts}, our study proposes using sparsely connected graph structures. This approach aims to reduce the series dimension redundancy by selectively connecting series that provide unique and informative insights. In addition, we employ global tokens to encapsulate and thereby mitigate the impact of redundancy within the time dimension. We tested our model on Low-rank datasets (see Section~\ref{lowRank}) with varying series numbers, highlighting its stable performance across increasing series dimensions, a key advantage over the less effective series-mixing methods in handling redundancy.}

Our contributions are threefold: First, we unveil the series-aware framework, a novel extension for Transformer-based models, utilizing global tokens to effectively exploit inter-series dependencies without adding undue complexity. Next, we introduce SageFormer, a series-aware Transformer tailored for long-term MTS forecasting. Unlike existing models, SageFormer excels in capturing both intra- and inter-series dependencies. Finally, extensive experiments on both real-world and synthetic datasets verify the effectiveness and superiority of SageFormer.


\section{Related Works}

\subsection{Multivariate Time Series Forecasting} MTS forecasting models can generally be categorized into statistical and deep models. Many forecasting methods begin with traditional tools such as the Vector Autoregressive model and the Vector Autoregressive Moving Average~\cite{Box1968SomeRA, box2015time}. These typical statistical MTS forecasting models assume linear dependencies between series and values. With the advancement of deep learning, various deep models have emerged and often demonstrate superior performance compared to their statistical counterparts. Temporal Convolutional Networks~\cite{BaiTCN2018, 2019Think} and DeepAR~\cite{Flunkert2017DeepARPF} consider MTS data as sequences of vectors, employing CNNs and RNNs to capture temporal dependencies. 

\subsection{Transformers for long-term MTS Forecasting} Recently, Transformer models with self-attention mechanisms have excelled in various fields~\cite{NIPS2017_3f5ee243, Devlin2019BERTPO, dong2018speech, liu2021Swin}. Numerous studies aim to enhance Transformers for long-term MTS forecasting by addressing their quadratic complexity. Notable approaches include Informer~\cite{haoyietal-informer-2021}, introducing ProbSparse self-attention and distilling techniques; Autoformer~\cite{wu2021autoformer}, incorporating decomposition and auto-correlation concepts; FEDformer~\cite{zhou2022fedformer}, employing a Fourier-enhanced structure; and Pyraformer~\cite{liu2021pyraformer}, implementing pyramidal attention modules. PatchTST~\cite{nie2022time} divides each series into patches and uses a series-independent Transformer to model temporal patterns. While these models primarily focus on reducing temporal dependencies modeling complexity, they often overlook crucial inter-series dependencies.

\subsection{Inter-series dependencies for MTS Forecasting} Several methods have been proposed to explicitly enhance inter-series dependencies in MTS forecasting. LSTnet~\cite{2018Modeling} employs CNN for inter-series dependencies and RNN for temporal dependencies. GNN-based models~\cite{li2017diffusion, yu2017spatio, cao2020spectral, wu2020connecting}, such as MTGNN~\cite{wu2020connecting}, utilize temporal and graph convolution layers to address both dependencies. STformer~\cite{grigsby2021long} flattens multivariate time series into a 1D sequence for Transformer input, while Crossformer~\cite{zhang2023crossformer} employs dimension-segment-wise embedding and a two-stage attention layer for efficient temporal and inter-series dependencies capture respectively. Although there exist several works showing the effectiveness of temporal convolutions to model long-range dependencies~\cite{gu2022efficiently, liu2022SCINet}, most traditional CNN and GNN-based methods still focus on short-term prediction and struggle to capture long-term temporal dependencies.
STformer~\cite{grigsby2021long} and Crossformer~\cite{zhang2023crossformer} extend 1-D attention to 2-D, but they fail to reveal the relationships between series explicitly. Unlike the methods mentioned above, our proposed framework serves as a general framework that can be applied to various Transformer-based models, utilizing graph structures to enhance their ability to represent inter-series dependencies.

\hlcyan{Recently, OneNet~\cite{zhang2023onenet} was introduced, employing dual models to separately address dependencies inter- and intra-series. While OneNet emphasizes online learning for short-term forecasts, our approach is different, focusing on offline learning for longer-term predictions. Unlike OneNet's separate treatment of inter- and intra-series dependencies, we adopt a joint modeling approach, to provide a more comprehensive understanding of time series dynamics. Our methodology seeks to enhance the predictive model's structure and functionality for improved long-term forecasts, distinguishing our work from OneNet's optimization-focused strategy.}

\section{Methodology}

\subsection{Problem Definition}
In this paper, we concentrate on long-term MTS forecasting tasks. Let $\mathbf{x}_t\in\mathbb{R}^C$ denote the value of $C$  series at time step $t$. Given a historical MTS sample $\mathbf{X}_t=\left[\mathbf{x}_t, \mathbf{x}_{t+1}, \cdots, \mathbf{x}_{t+H-1}\right] \in\mathbb{R}^{C\times H}$ with length $H$, the objective is to predict the next $T$ steps of MTS values $\mathbf{Y}_t=\left[\mathbf{x}_{t+L}, \cdots, \mathbf{x}_{t+L+T-1}\right] \in\mathbb{R}^{C\times T}$. The aim is to learn a mapping $f(\cdot): \mathbf{X}_t \rightarrow \mathbf{Y}_t$ using the proposed model (we omit the subscript $t$ when it does not cause ambiguity).

We employ graphs to represent inter-series dependencies in MTS and briefly overview relevant graph-related concepts. From a graph perspective, different series in MTS are considered nodes, and relationships among series are described using the graph adjacency matrix. Formally, the MTS data can be viewed as a signal set $\mathcal{G}=\left\{\mathcal{V}, \mathbf{X}_t, \mathbf{A}\right\}$. The node set $\mathcal{V}$ contains $C$ series of MTS data and $\mathbf{A}\in \mathbb{R}^{C\times C}$ is a weighted adjacency matrix. The entry $a_{ij}$ indicates the dependencies between series $i$ and $j$. If they are not dependent, $a_{ij}$ equals zero. The main symbols used in the paper and their meanings are detailed in Table~\ref{tab:notations}.

\subsection{Overview}

The series-aware framework, as illustrated in Algorithm~\ref{alg}, is designed to forecast MTS data. The workflow mainly consists of two components: the generation of global tokens, and an iterative message passing procedure. Within the message passing mechanism, contemporary architectures can be utilized for both inter-series and intra-series information propagation.

SageFormer is an instantiation of the Series-Aware framework. It effectively employs GNNs to model inter-series dependencies, while leveraging Transformers to capture intra-series dependencies. Thus, it ensures the model comprehensively grasps the underlying dynamics of the series-aware setup. The overall structure adheres to a Transformer encoder design, and replaces the conventional Transformer decoder with a more efficient linear decoder head ($\operatorname{ForecastingHead}$), as supported by the literature~\cite{Zeng2022AreTE, nie2022time, wu2023timesnet}. With its unique combination of GNNs and Transformers, SageFormer emerges as a promising solution for capturing and modeling the essence of intra- and inter-series relations.

\begin{table}[th]
\centering
\caption{Table of Notations}\label{tab:notations}
\begin{tabular}{cp{5cm}}
\hline
\textbf{Symbol} & \textbf{Description} \\
\hline
$\mathrm{C}$ & Number of series in MTS data \\
$H$ & Historical input length \\
$T$ & Predictive horizontal length \\
$\mathbf{x}_t\in\mathbb{R}^{C}$ & The value of $C$ series at time step $t$ \\
\hlcyan{$\mathbf{X}_t\in\mathbb{R}^{C\times H}$} & \hlcyan{A historical MTS sample with length $H$} \\
$\mathbf{Y}_t\in\mathbb{R}^{C\times T}$ & The next $T$ steps of MTS values to be predicted \\
$f(\cdot)$ & Mapping function from $\mathbf{X}_t$ to $\mathbf{Y}_t$ \\
$\mathcal{G}=\left\{\mathcal{V}, \mathbf{X}_t, \mathbf{A}\right\}$ & MTS data viewed from a graph perspective \\
$\mathcal{V}$ & Node set containing $C$ series of MTS data \\
$\mathbf{A}\in\mathbb{R}^{C\times C}$ & Weighted adjacency matrix \\
$a_{ij}$ & Entry indicating the dependencies between series $i$ and $j$ \\
\hlcyan{$\mathcal{X}^{(l)}\in\mathbb{R}^{C\times(M+N)\times D_h}$} & \hlcyan{Embedding with global tokens of $l$-th layer} \\
$L$ & Number of encoder layers \\
$\mathrm{TEB}$ & Transformer Encoding Block \\
$\mathrm{GNN}$ & Graph Neural Network \\
$\operatorname{ForecastingHead}$ & Decoder head that produces the predictions \\
\hline
\end{tabular}
\end{table}

\begin{algorithm}[t]
\caption{Series-aware Framework's Workflow}
\label{alg}
\begin{algorithmic}[1] 
\Require The input MTS history $\mathbf{X}$
\Ensure The predicted MTS future $\mathbf{Y}$

\State $\mathcal{X}^{(0)} \leftarrow \operatorname{GlobalEembedding}(\mathbf{X})$ \Comment{global tokens}

\For{$c=1, \dots, C$}
    \State $\mathcal{X}^{(1)}_{c,:,:} \leftarrow \mathrm{IntraSeries}\left(\mathcal{X}^{(0)}_{c,:,:}\right)$
\EndFor

\For{$l = 2, \dots, L-1$}\Comment{iterative message passing} 
    \For{$m=1, ..., M$}
        \State $\widehat{\mathcal{X}}^{(l)}_{:,m,:} \leftarrow \mathrm{InterSeries}\left(\mathcal{X}^{(l)}_{:,m,:} \right)$ \Comment{series-aware}
    \EndFor
    \For{$c=1, \dots, C$}
        \State $\mathcal{X}^{(l+1)}_{c,:,:} \leftarrow \mathrm{IntraSeries}\left(\widehat{\mathcal{X}}^{(l)}_{c,:,:}\right)$ \Comment{temporal}
    \EndFor
\EndFor

\State $\hat{\mathcal{X}} \leftarrow \left\{ \mathcal{X}^{(L)}_{m,:,:} | m>M\right\}$ \\
\Return $Y \leftarrow \mathrm{ForecastingHead}\left(\hat{\mathcal{X}} \right)$

\end{algorithmic}
\end{algorithm}

\begin{figure*}[t]
\centering
     \includegraphics[width=\textwidth]{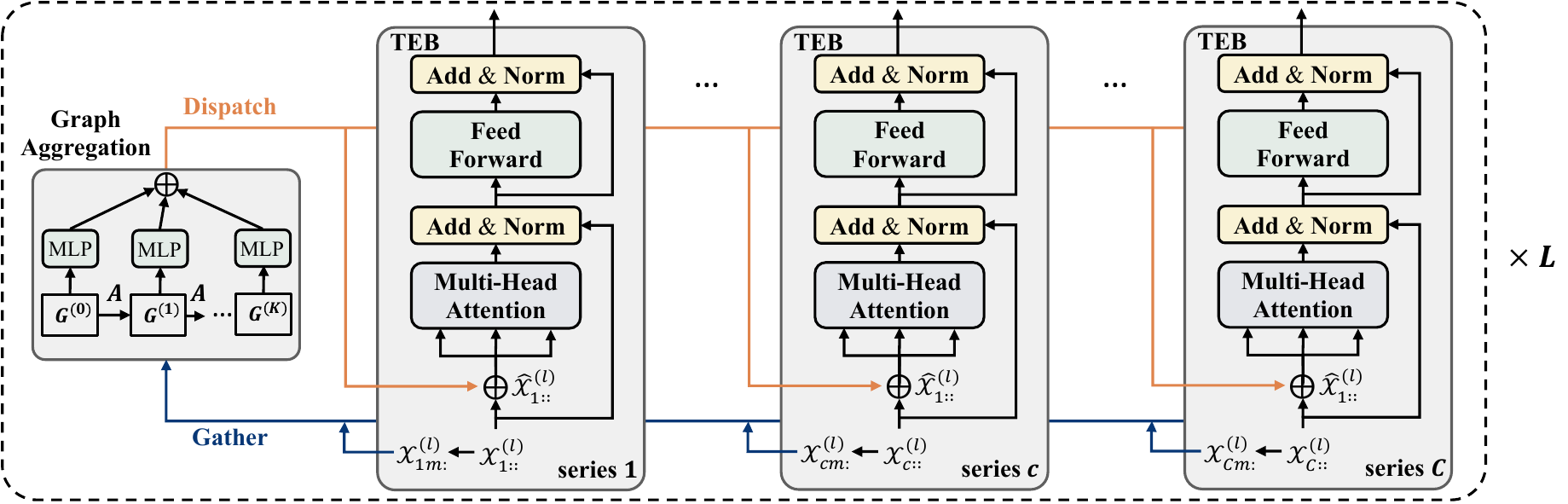}
      \caption{Illustration of the iterative message-passing process in SageFormer. Each layer begins with graph aggregation, where global tokens from all series are gathered and processed by the multi-hop GNN component (leftmost rectangle). Graph-enhanced global tokens are then dispatched to their original series and encoded by TEB. The weights of each TEB are shared among all series.}
\label{fig:SageFormerModel}
\end{figure*}

\subsection{Series-aware Global Tokens} \label{series-aware_embedddings}

The traditional approach in Transformers involves obtaining input tokens by projecting point-wise or patch-wise split input time-series~\cite{haoyietal-informer-2021,zhou2022fedformer,nie2022time}. This is executed with the intention that these tokens embody local semantic information, which can then be examined for their mutual connections through self-attention mechanisms.


\hlcyan{Our methodology introduces a pivotal innovation: the integration of global tokens to enhance series awareness, a concept inspired by the class token in Natural Language Processing models~\cite{Devlin2019BERTPO} and Vision Transformer~\cite{dosovitskiy2021an}. We prepend randomly initialized learnable tokens for each series to encapsulate their corresponding global information. These global tokens are not just placeholders; they are a distinctive aspect of our approach, functioning to capture and consolidate intra-series temporal information after the first layer of self-attention. As detailed in Section~\ref{sec:iterative}, we leverage these global tokens to capture inter-series dependencies effectively. This is achieved through their involvement in GNN-based inter-series information transfer, thereby augmenting the series awareness of each sub-series (for further discussion, please see Section~\ref{sec:dis_on_global_tokens}).
}

In accordance with PatchTST~\cite{nie2022time}, the input MTS $\mathbf{X} \in \mathbb{R}^{C\times H}$ is reshaped into overlapped patches $\mathcal{X}_p=\{\mathbf{X}_1, \cdots, \mathbf{X}_C\} \in \mathbb{R}^{C\times N\times P}$. Here P is the subsequence patch length, and $N=\left\lfloor(H-P)/S\right\rfloor+2$ denotes the number of patches. Additionally, $S$ indicates the non-overlapping length between adjacent patches. $\mathbf{X}_c=\{\mathbf{x}_c^1, \cdots,\mathbf{x}_c^N\}\in\mathbb{R}^{N\times P}$ represents the patched sequence for series $c$. To facilitate the processing within the Transformer encoding blocks (TEB), each patch in \(\mathcal{X}_p\) is projected into a consistent latent space of size \(D_h\) using a trainable linear projection (\(\mathbf{E}_{proj}\in \mathbb{R}^{P\times D_h}\)), as detailed in Equation~\ref{eq1}.

\hlcyan{$M$ learnable global tokens (randomly initialized using Gaussian distribution) $\mathbf{g}_i\in\mathbb{R}^{D_h}$ are added before the patched sequences, representing each series' global information enhanced by self-attention, resulting in an effective input token sequence length of $M+N$.} Positional information is enhanced via 1D learnable additive position embeddings $\mathbf{E}_{pos}$. The final embedding of input MTS sample $\mathbf{X}$ is $\mathcal{X}^{(0)}\in \mathbb{R}^{C\times (M+N)\times D_h}$, where
\begin{equation}
    \label{eq1}
    \mathcal{X}^{(0)}_{c,:,:}=\left[\mathbf{g}_{1};\cdots; \mathbf{g}_{M};\mathbf{x}^1_c\mathbf{E}_{proj};\cdots;\mathbf{x}^{N}_{c}\mathbf{E}_{proj}\right]+\mathbf{E}_{pos}.
\end{equation}

\subsection{Graph Structure Learning}
In SageFormer, the adjacency matrix is learned end-to-end, enabling the capture of relationships across series without the necessity of prior knowledge. In MTS forecasting, we posit that inter-series dependencies are inherently unidirectional. For example, while the power load may influence oil temperature, the converse isn't necessarily valid. Such directed relationships are represented within the graph structure we derive. It's worth noting that many time series lack an intrinsic graph structure or supplementary auxiliary data. However, our methodology is adept at deducing the graph structure solely from the data, obviating the need for external inputs. This attribute enhances its versatility and wide-ranging applicability.

\begin{table*}[hbt]
    \caption{The statistics of the six mainstream datasets.}\label{tab:datasets}
    \centering
    \begin{tabular}{c|ccccccc} \hline
    Datasets & Traffic & Electricity & Weather & Exchange & ILI & ETTh1/ETTh2 & ETTm1/ETTm2 \Bstrut \Tstrut \\ 
    \hline
    Number of Series & 862 & 321 & 21 & 8 & 7 & 7 & 7 \Tstrut \\
    Timesteps & 17,544 & 26,304 & 52,696 & 7588 & 966 & 17,420 & 69,680 \\
    Temporal Granularity & 1 hour & 1 hour & 10 mins & 1 day & 1 week  & 1 hour & 15 mins \Bstrut \\
    \hline
    \end{tabular}
\end{table*}

Specifically, the entire graph structure learning module can be described by the following equations:
\begin{align}
    \mathbf{M}_1 &= \operatorname{act}(\mathbf{E}\mathbf{\Theta}_1); \ \mathbf{M}_2 = \operatorname{act}(\mathbf{E}\mathbf{\Theta}_2) \label{nodeEmb} \\
    \mathbf{A}' &=\operatorname{Relu}(\mathbf{M}_1\mathbf{M}_2^T-\mathbf{M}_2\mathbf{M}_1^T) \label{uni}
\end{align}
Node embeddings of series are learned through randomly initialized $\mathbf{E}\in \mathbb{R}^{C\times D_g}$, where $D_g$ is the number of features in node embeddings. Subsequently, $\mathbf{E}$ is transformed into $\mathbf{M}\in\mathbb{R}^{C\times D_g}$ using trainable parameters $\mathbf{\Theta_1}, \mathbf{\Theta_2}\in\mathbb{R}^{D_g\times D_g}$, with the nonlinear activation function $\operatorname{act}$ (Equation~\ref{nodeEmb}). Following the approach of MTGNN~\cite{wu2020connecting}, Equation~\ref{uni} is employed to learn unidirectional dependencies. For each node, the top 
$K$ nearest nodes are identified as neighbors from $\mathbf{A}'$. Weights of nodes that aren't connected are set to zero. This process results in the final sparse adjacency matrix $\mathbf{A}\in\mathbb{R}^{C\times C}$ of the uni-directional graph $\mathcal{G}$.

\subsection{Iterative Message Passing} \label{sec:iterative}
The embedding tokens (outlined in Section~\ref{series-aware_embedddings}) are processed by SageFormer encoder layers, where temporal encoding and graph aggregation are conducted iteratively (Fig.~\ref{fig:SageFormerModel}). This approach aims to disseminate the global information gathered during the GNN phase among all tokens within each series. As a result, the model captures both intra- and inter-series dependencies through iterative message passing.

\paragraph{Graph Aggregation} The graph aggregation phase aims to fuse each series' information with its neighbors' information, thereby enhancing each series' representation with related patterns. For each series in the $l$-th layer, we take the first $M$ embeddings as the global tokens of layer $l$: $\mathbf{G}_{i}^{(l)} \leftarrow \mathcal{X}^{(l)}_{:,i,:} \in \mathbb{R}^{C\times D_h},\, i\leq M$. The global tokens of layer $l$ are gathered from all series and passed into the GNN for graph aggregation. For simplicity, we employ the same model as \cite{li2017diffusion, wu2020connecting} for graph aggregation:
\begin{equation}
     \widehat{\mathbf{G}}_{i} = \sum_{d=0}^D \tilde{\mathbf{A}}^d \mathbf{G}_i \mathbf{W}_{d}, \ i\leq M \label{kpopFuse}
\end{equation}
Equation~\ref{kpopFuse} represents multi-hop information fusion on the graph, where $D$ denotes the depth of graph aggregation and $\tilde{\mathbf{A}}$ is the graph Laplacian matrix. \hlcyan{The term $\mathbf{W}_{d}\in \mathbb{R}^{D\times D}$ is a set of learnable parameters, specifically a series of weight matrices corresponding to each depth level $d$. These matrices transform and scale the feature vectors at each layer of the graph, allowing the model to capture and integrate information from different neighborhood scales effectively.} Each of the embeddings $\widehat{\mathbf{G}}_{i}$ is dispatched to its original series and then concatenated with series tokens, resulting in graph-enhanced embeddings $\widehat{\mathcal{X}}^{(l)}$.

\paragraph{Temporal Encoding} The graph-enhanced embeddings are later processed by Transformer components (Transformer~\cite{NIPS2017_3f5ee243}, Informer~\cite{haoyietal-informer-2021}, FEDformer~\cite{zhou2022fedformer}, etc.). We choose the vanilla Transformer encoder blocks~\cite{NIPS2017_3f5ee243} as our backbone. The output of the TEB functions as input token-level embeddings for the following encoding layer. Previously aggregated information from the GNN is disseminated to other tokens within each series via self-attention, enabling access to related series information. This process enhances the expressiveness of our model compared to series-independent models.

\hlcyan{Note that both the node embeddings and global tokens are randomly initialized and subsequently optimized through an iterative message passing process. This approach, while unique in its emphasis on tokens, aligns with typical GNN methods in graph aggregation. The input tensors for both approaches share the same format, characterized by the dimensions $(b, n, d)$, where $b$ represents the batch size, $n$ the number of nodes in the graph, and $d$ the feature dimension. Additionally, due to the incorporation of a temporal encoding operation at the initial layer, the global tokens, during the graph aggregation phase, are not mere random entities. Instead, they embody comprehensive global information within the variables. This design ensures that the global tokens positively contribute to the model's convergence.}



\section{Experiments}
\subsection{Experimental Setup}
\paragraph{Datasets} To evaluate our proposed series-aware framework and SageFormer, extensive experiments have been conducted on six mainstream real-world datasets, including Weather, Traffic, Electricity, ILI(Influenza-Like Illness), and four ETT(Electricity Transformer Temperature) datasets and two synthetic datasets (see Section~\ref{sec:synthetic}). The statistics of these multivariate datasets are shown in Table~\ref{tab:datasets}. Among all datasets, Traffic and Electricity have more series, which can better reflect the effectiveness of the proposed method. The details are described below:

\begin{enumerate}
\item The Traffic dataset\footnote{\url{https://pems.dot.ca.gov/}} contains data from 862 sensors installed on highways in the San Francisco Bay Area, measuring road occupancy. This dataset is recorded on an hourly basis over two years, resulting in a total of 17,544 time steps.
\item The Electricity Consumption Load (Electricity) dataset\footnote{\url{https://archive.ics.uci.edu/ml/datasets/ElectricityLoadDiagrams20112014}} records the power usage of 321 customers. The data is logged hourly over a two-year period, amassing a total of 26,304 time steps.
\item The Weather dataset\footnote{\url{https://www.bgc-jena.mpg.de/wetter}} is a meteorological collection featuring 21 variates, gathered in the United States over a four-year span.
\item The Exchange dataset\footnote{\url{https://github.com/laiguokun/multivariate-time-series-data}} details the daily foreign exchange rates of eight different countries, including Australia, British, Canada, Switzerland, China, Japan, New Zealand and Singapore ranging from 1990 to 2016.
\item The Influenza-Like Illness (ILI) dataset\footnote{\url{https://gis.cdc.gov/grasp/fluview/fluportaldashboard.html}} is maintained by the United States Centers for Disease Control and Prevention. It collates patient information on a weekly basis for a period spanning from 2002 to 2021.
\item The Electricity Transformers Temperature (ETT) datasets\footnote{\url{https://github.com/zhouhaoyi/ETDataset}} are procured from two electricity substations over two years. They provide a summary of load and oil temperature data across seven variates. For ETTm1 and ETTm2, the "m" signifies that data was recorded every 15 minutes, yielding a total of 69,680 time steps. ETTh1 and ETTh2 represent the hourly equivalents of ETTm1 and ETTm2, respectively, each containing 17,420 time steps.
\end{enumerate}

\begin{table*}[th]
  \caption{\hlcyan{Long-term Forecasting Task Results. See Table \ref{tab:full_forecasting_results} in Appendix For the Full Results. Superscript \textit{a} marks the models explicitly utilizing inter-series dependencies; \textit{b} marks series-independent neural models; \textit{c} marks series-mixing transformer-based models. }}\label{tab:long_term_forecasting_results_short}
  \centering
  \begin{threeparttable}
  \renewcommand{\multirowsetup}{\centering}
  \setlength{\tabcolsep}{2.6pt}
  \begin{tabular}{c|cc|cc|cc|cc|cc|cc|cc|cc|cc|cc}
    \toprule
  \multicolumn{1}{c}{\multirow{2}{*}{Models}} & 
  \multicolumn{2}{c}{\rotatebox{0}{\cellcolor{red!25}\textbf{SageFormer}}} & 
  \multicolumn{2}{c}{\rotatebox{0}{\cellcolor{blue!25}\scalebox{0.92}{Crossformer\tnote{a}}}} & 
  \multicolumn{2}{c}{\rotatebox{0}{\cellcolor{blue!25}MTGNN\tnote{a}}} & 
  \multicolumn{2}{c}{\rotatebox{0}{\cellcolor{blue!25}LSTnet\tnote{a}}} & 
  \multicolumn{2}{c}{\rotatebox{0}{\cellcolor{green!25}PatchTST\tnote{b}}} & 
  \multicolumn{2}{c}{\rotatebox{0}{\cellcolor{green!25}DLinear\tnote{b}}} & 
  \multicolumn{2}{c}{\rotatebox{0}{\cellcolor{yellow!25}Stationary\tnote{c}}} & 
  \multicolumn{2}{c}{\rotatebox{0}{\cellcolor{yellow!25}Autoformer\tnote{c}}} & 
  \multicolumn{2}{c}{\rotatebox{0}{\cellcolor{yellow!25}Informer\tnote{c}}} & 
  \multicolumn{2}{c}{\rotatebox{0}{\cellcolor{yellow!25}Transformer\tnote{c}}}  
    \\
    \multicolumn{1}{c}{} & 
    \multicolumn{2}{c}{\cellcolor{red!25}(\textbf{Ours})} & 
    \multicolumn{2}{c}{\cellcolor{blue!25}\cite{zhang2023crossformer}} &
    \multicolumn{2}{c}{\cellcolor{blue!25}\cite{wu2020connecting}} &
    \multicolumn{2}{c}{\cellcolor{blue!25}\cite{2018Modeling}} &
    \multicolumn{2}{c}{\cellcolor{green!25}\cite{nie2022time}} &
    \multicolumn{2}{c}{\cellcolor{green!25}\cite{Zeng2022AreTE}} &
    \multicolumn{2}{c}{\cellcolor{yellow!25}\cite{Liu2022NonstationaryTR}} &
    \multicolumn{2}{c}{\cellcolor{yellow!25}\cite{wu2021autoformer}} & 
    \multicolumn{2}{c}{\cellcolor{yellow!25}\cite{haoyietal-informer-2021}} &
    \multicolumn{2}{c}{\cellcolor{yellow!25}\cite{NIPS2017_3f5ee243}} 
    \\
    \cmidrule(lr){2-3} \cmidrule(lr){4-5}\cmidrule(lr){6-7} \cmidrule(lr){8-9}\cmidrule(lr){10-11}\cmidrule(lr){12-13}\cmidrule(lr){14-15}\cmidrule(lr){16-17}\cmidrule(lr){18-19}\cmidrule(lr){20-21}
    \multicolumn{1}{c}{Metric} & MSE & MAE & MSE & MAE & MSE & MAE & MSE & MAE & MSE & MAE & MSE & MAE & MSE & MAE & MSE & MAE & MSE & MAE & MSE & MAE  \\
    \toprule 
    Traffic&\boldres{0.436}&\boldres{0.285}&0.570&0.312&0.592&0.317&0.736&0.450&\secondres{0.472}&\secondres{0.298}&0.625&0.383&0.624&0.340&0.628&0.379&0.764&0.416&0.661&0.363\\
    \midrule
    Electricity &\boldres{0.175}&\boldres{0.273}&0.314&0.366&0.333&0.378&0.440&0.494&0.200&\secondres{0.288}&0.212&0.300&\secondres{0.193}&0.296&0.227&0.338&0.311&0.397&0.272&0.367\\
	  \midrule
    Weather &\boldres{0.247}&\boldres{0.273}&\secondres{0.256}&0.305&0.290&0.348&0.768&0.672&\secondres{0.256}&\secondres{0.279}&0.265&0.317&0.288&0.314&0.338&0.382&0.634&0.548&0.611&0.557\\
	  \midrule
    ETTm1 &\boldres{0.388}&\boldres{0.400}&0.509&0.507&0.566&0.537&1.947&1.206&\secondres{0.389}&\secondres{0.401}&0.403&0.407&0.481&0.456&0.588&0.517&0.961&0.734&0.936&0.728\\
	  \midrule
    ETTm2 &\boldres{0.277}&\boldres{0.322}&1.433&0.747&1.287&0.751&2.639&1.280&\secondres{0.280}&\secondres{0.328}&0.350&0.401&0.306&0.347&0.327&0.371&1.410&0.810&1.478&0.873\\
	  \midrule
    ETTh1 &\boldres{0.431}&\boldres{0.433}&0.615&0.563&0.679&0.605&2.113&1.237&\secondres{0.443}&\secondres{0.443}&0.456&0.452&0.570&0.537&0.496&0.487&1.040&0.795&0.919&0.759\\
	  \midrule
    ETTh2 &\boldres{0.374}&\boldres{0.403}&2.170&1.175&2.618&1.308&4.382&2.008&\secondres{0.381}&\secondres{0.404}&0.559&0.515&0.526&0.516&0.450&0.459&4.431&1.729&4.492&1.691\\
	  \midrule
    Exchange &\boldres{0.354}&\boldres{0.399}&0.756&0.645&0.786&0.674&1.681&1.197&\boldres{0.354}&\secondres{0.400}&\boldres{0.354}&0.414&0.461&0.454&0.613&0.539&1.550&0.998&1.386&0.898\\
	  \midrule
    ILI &2.113&\boldres{0.877}&3.417&1.215&4.861&1.507&5.300&1.657&\boldres{2.065}&\secondres{0.882}&2.616&1.090&\secondres{2.077}&0.914&3.006&1.161&5.137&1.544&4.784&1.471\\
  \bottomrule
  \end{tabular}
      \begin{tablenotes}
          \item[1] 
          \textbf{Bold}/\underline{underline} indicates the \textbf{best}/\underline{second}. The results is averaged from all four prediction lengths (96, 192, 336, 720). 
	\end{tablenotes}
  \end{threeparttable}
\end{table*}

\paragraph{Baselines and Task Settings} We compare our proposed method with nine popular models for long-term MTS forecasting problems as baselines, including three models that explicitly utilize inter-series dependencies: Crossformer~\cite{zhang2023crossformer}, MTGNN~\cite{wu2020connecting}, and LSTnet~\cite{2018Modeling}; two series-independent neural models: DLinear~\cite{Zeng2022AreTE} and PatchTST~\cite{nie2022time}; and four series-mixing transformer-based models: Transformer~\cite{NIPS2017_3f5ee243}, Informer~\cite{haoyietal-informer-2021}, Autoformer~\cite{wu2021autoformer}, and Non-stationary Transformer~\cite{Liu2022NonstationaryTR}.  \label{sec:task_settings}

All baselines are reproduced using the original paper's configuration or the official code. However, the only exception is that we standardize the look-back window across all models to 36 for the ILI dataset and 96 for the remaining datasets to ensure a fair comparison. Consequently, some discrepancies may exist between our input-output setting and those reported in the referenced papers. We exclude traditional time series forecasting models (such as ARIMA, LSTM) from our baselines, as Transformer-based models have been demonstrated to outperform these in long-term forecasting tasks~\cite{haoyietal-informer-2021}.

\paragraph{Implementation details} For model training and evaluation, we adopt the same settings as in~\cite{wu2023timesnet}. \hlcyan{The dataset is processed with a stride of one, creating various input-output pairs. Train/Val/Test sets are zero-mean normalized using the mean and standard deviation of the training set. The historical sequence length is set as 36 for ILI and 96 for the others. Performance is evaluated over varying future window sizes on each dataset.  Mean Square Error (MSE) and Mean Absolute Error (MAE) serve as evaluation metrics. To ensure reliability, each experiment is conducted five times on an NVIDIA TITAN RTX 3090 24GB GPU, averaging the results to counteract random variations.}

\hlcyan{SageFormer employs a time-based optimization strategy, iterating over the dataset's time axis with a stride of 1. The graph structure $A$ is constructed during training and remains fixed during testing. Despite incorporating both GNN and Transformer components, all weights are shared and optimized together using the Mean Squared Error (MSE) criterion. The optimization objective is formulated as
$\min_{\theta} \frac{1}{N}\sum_{i=1}^{N}(\mathbf{Y}_i - f(\mathbf{X}_i;\theta))^2$
, where $N$ is the total number of training samples and $\theta$ embodies the trainable parameters of the model.
Optimization is conducted via the Adam optimizer with a learning rate of 1e-4, and the model is trained for approximately 20 epochs with an early stop mechanism to prevent overfitting.}

By default, SageFormer comprises two encoder layers, an attention head count ($H$) of 16, and a model hidden dimension ($D_h$) of 512. We designate the number of global tokens as $M=1$; the node embedding dimension $D_g$ is set to 16, the number of nearest neighbors is $K=16$, and the graph aggregation depth is $D=3$. Furthermore, we employ a patch length ($P$) of 24 and a stride ($S$) of 8, consistent with the parameters utilized in PatchTST. For the larger datasets (Traffic, Electricity), a model with three encoder layers is adopted to enhance the expressive capacity.

\subsection{Main Results}

\begin{figure}[t!]
\centering
\includegraphics[width=\columnwidth]{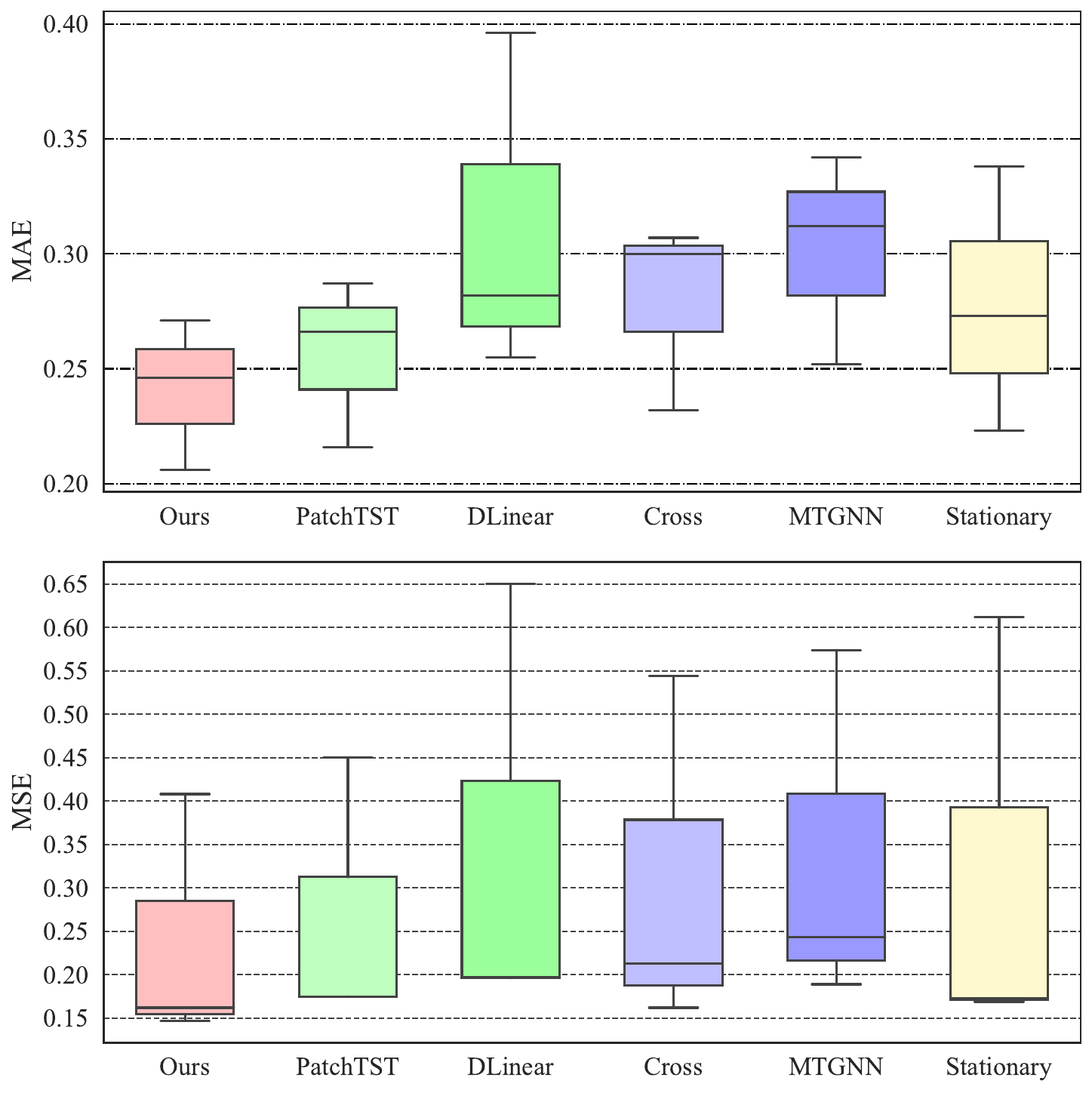}
\caption{\hlcyan{MAE / MSE results on Traffic, Electricity, and Weather datasets.}}
\label{fig:boxplot}
\end{figure}

\begin{table*}[th]
	\centering
	\caption{\hlcyan{Performance Promotion by Applying the Proposed Framework to Transformer and Its Variants.}}\label{tab:add-on-transformer_short}

   \begin{threeparttable}
    \begin{tabular}{c|cc|cc|cc|cc|cc|cc} 
    \toprule
    \multicolumn{1}{c}{Dataset} & \multicolumn{2}{c}{Transformer} & \multicolumn{2}{c}{\textbf{+ Series-aware} \tnote{2}} & \multicolumn{2}{c}{Informer \tnote{3}} & \multicolumn{2}{c}{\textbf{+ Series-aware}} & \multicolumn{2}{c}{FEDformer} & \multicolumn{2}{c}{\textbf{+ Series-aware}}\\  
    \cmidrule(lr){2-3} \cmidrule(lr){4-5} \cmidrule(lr){6-7} \cmidrule(lr){8-9} \cmidrule(lr){10-11} \cmidrule(lr){12-13}
    \multicolumn{1}{c}{Metric} & MSE & MAE & MSE & MAE & MSE & MAE & MSE & MAE & MSE & MAE & MSE & MAE\\ 
    \toprule
    \rotatebox{0}{\scalebox{0.95}{Traffic}}
    & 0.661 & 0.362 & \boldres{0.549} & \boldres{0.293} & 0.610 & 0.376 & \boldres{0.578} & \boldres{0.318} & 0.764 & 0.416 & \boldres{0.707} & \boldres{0.396} \\ 
    \midrule

    \rotatebox{0}{\scalebox{0.95}{Electricity}}
    & 0.272 & 0.367 & \boldres{0.202} & \boldres{0.290} & 0.214 & 0.327 & \boldres{0.207} & \boldres{0.305} & 0.311 & 0.397 & \boldres{0.223} & \boldres{0.319} \\ 
    \midrule
      
    \rotatebox{0}{\scalebox{0.95}{Weather}}
    & 0.611 & 0.557 & \boldres{0.290} & \boldres{0.348} & 0.309 & 0.360 & \boldres{0.285} & \boldres{0.323} & 0.634 & 0.548 & \boldres{0.269} & \boldres{0.304} \\ 
    \midrule

    \rotatebox{0}{\scalebox{0.95}{ETTh1}}
    & 0.919 & 0.759 & \boldres{0.459} & \boldres{0.456} & 0.440 & 0.460 & \boldres{0.433} & \boldres{0.442} & 1.040 & 0.795 & \boldres{0.673} & \boldres{0.577} \\ 
    \bottomrule
    \end{tabular}

    \begin{tablenotes}
        \item[1] The inconsistency between the Informer results in Table \ref{tab:long_term_forecasting_results_short} and Table \ref{tab:add-on-transformer_short} arises because of different input history lengths.
    \end{tablenotes}
  \end{threeparttable}

\end{table*}

\begin{table*}[htbp]
  \caption{Model Architecture Ablations. }\label{tab:ablation_on_architecture}
  \centering
  \begin{threeparttable}

  \renewcommand{\multirowsetup}{\centering}
  \begin{tabular}{c|cccc|c|cccc|c}
    \toprule
    \multicolumn{1}{c}{Datasets} & 
    \multicolumn{4}{c}{Traffic} &
    \multicolumn{1}{c}{\multirow{2}{*}{AVG}} &
    \multicolumn{4}{c}{ETTh1} &
    \multicolumn{1}{c}{\multirow{2}{*}{AVG}} \\
    \cmidrule(lr){2-5} \cmidrule(lr){7-10} 
    Prediction Lengths & 96 & 192 & 336 & 720 & {} & 96 & 192 & 336 & 720\\
    \toprule
        \scalebox{1.00}{SageFormer}
        & \scalebox{1.00}{\boldres{0.408}} & \scalebox{1.00}{\boldres{0.421}} & \scalebox{1.00}{\boldres{0.438}} & \scalebox{1.00}{\boldres{0.477}} 
        & \scalebox{1.00}{\boldres{0.436}}
        & \scalebox{1.00}{\secondres{0.377}} & \scalebox{1.00}{\boldres{0.423}} & \scalebox{1.00}{\boldres{0.459}} & \scalebox{1.00}{\boldres{0.465}} 
        & \scalebox{1.00}{\boldres{0.431}} \Bstrut \\
        \hline
        
        \scalebox{1.00}{- Graph Aggregation}
        & \scalebox{1.00}{0.446} & \scalebox{1.00}{0.452} & \scalebox{1.00}{0.466} & \scalebox{1.00}{0.506} 
        & \scalebox{1.00}{0.468}
        & \scalebox{1.00}{\boldres{0.372}} & \scalebox{1.00}{0.439} & \scalebox{1.00}{0.468} & \scalebox{1.00}{0.491}
        & \scalebox{1.00}{0.443} \Tstrut \\ %

        \scalebox{1.00}{- Global Tokens}
        & \scalebox{1.00}{0.420} & \scalebox{1.00}{\secondres{0.440}} & \scalebox{1.00}{0.457} & \scalebox{1.00}{0.507}
        & \scalebox{1.00}{0.456}
        & \scalebox{1.00}{0.381} & \scalebox{1.00}{0.431} & \scalebox{1.00}{0.462} & \scalebox{1.00}{0.478}
        & \scalebox{1.00}{0.438} \\ %
        
        \scalebox{1.00}{- Sparse Graph}
        & \scalebox{1.00}{0.416} & \scalebox{1.00}{0.441} & \scalebox{1.00}{\secondres{0.445}} & \scalebox{1.00}{\secondres{0.484}} 
        & \scalebox{1.00}{\secondres{0.447}}
        & \scalebox{1.00}{0.377} & \scalebox{1.00}{\secondres{0.423}} & \scalebox{1.00}{\secondres{0.459}} & \scalebox{1.00}{\secondres{0.467}}
        & \scalebox{1.00}{\secondres{0.432}} \\ %
        
        \scalebox{1.00}{- Directed Graph}
        & \scalebox{1.00}{\secondres{0.415}} & \scalebox{1.00}{0.442} & \scalebox{1.00}{0.449} & \scalebox{1.00}{0.494} 
        & \scalebox{1.00}{0.450}
        & \scalebox{1.00}{0.379} & \scalebox{1.00}{0.424} & \scalebox{1.00}{0.461} & \scalebox{1.00}{0.468}
        & \scalebox{1.00}{0.433} \\ %
        
        \bottomrule
    \end{tabular}

    \begin{tablenotes}
        \item[1] MSE metrics are reported. \emph{-Graph Aggregation}: Elimination of graph aggregation; \emph{-Global Tokens}: Graph information propagation applied to all tokens; \emph{-Sparse Graph}: Removal of the k-nearest neighbor constraint in graph learning; \emph{-Directed Graph}: Modification of graph learning from directed to undirected graphs.
    \end{tablenotes}
  \end{threeparttable}
\end{table*}

\paragraph{Long-term forecasting results}
\hlcyan{Table~\ref{tab:long_term_forecasting_results_short} presents the averaged forecasting results for the proposed SageFormer and other baseline models (see Table~\ref{tab:full_forecasting_results} in Appendix for the full results).} Overall, the forecasting results presented distinctly highlight the superior performance of our proposed SageFormer when compared to other baseline models. This superior performance is consistently observed across all benchmarks and prediction lengths. Even when juxtaposed against models that explicitly model inter-series dependencies, SageFormer emerges as the more effective solution. This suggests that our method possesses a robust capability to discern and capitalize on the relationships that exist among multiple series.

Digging deeper into the results, it becomes evident that SageFormer's prowess is even more pronounced on datasets with a higher number of variables. Specifically, on datasets such as Traffic, which has 862 predictive variables, and Electricity, with 321 predictive variables, our method offers significant enhancements. The improvements on previous state-of-the-art results are noteworthy – a 7.4\% average MSE reduction for Traffic (from $0.471$ to $0.436$) and a 9.3\% average MSE reduction for Electricity (from $0.193$ to $0.175$). To further elucidate this point, we've visualized the results of SageFormer against strong baselines for three relatively large datasets using box plots in Fig.~\ref{fig:boxplot}. These plots reveal that SageFormer not only attains a lower error mean but also demonstrates a smaller variance compared to other models, underscoring its consistent and reliable forecasting capabilities.


\paragraph{Framework generality} Furthermore, our model serves as a versatile extension for Transformer-based architectures. To validate this, we apply the SageFormer framework to three prominent Transformers and report the performance enhancement of each model as Table~\ref{tab:add-on-transformer_short}. Our method consistently improves the forecasting ability of different models, demonstrating that the proposed series-aware framework is an effective, universally applicable framework. By leveraging graph structures, it can better utilize the intra- and inter-series dependencies, ultimately achieving superior predictive performance.

\begin{figure*}[t]
\centering
\subfloat[]{\label{fig:hyper_global_token} \includegraphics[width=0.24\textwidth]{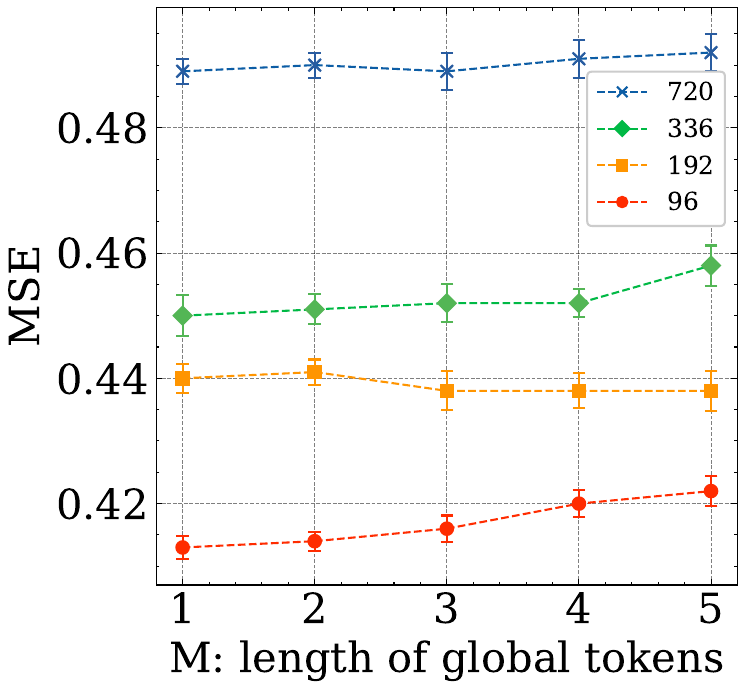}}
\hfill
\subfloat[]{\label{fig:hyper_graph_depth} \includegraphics[width=0.24\textwidth]{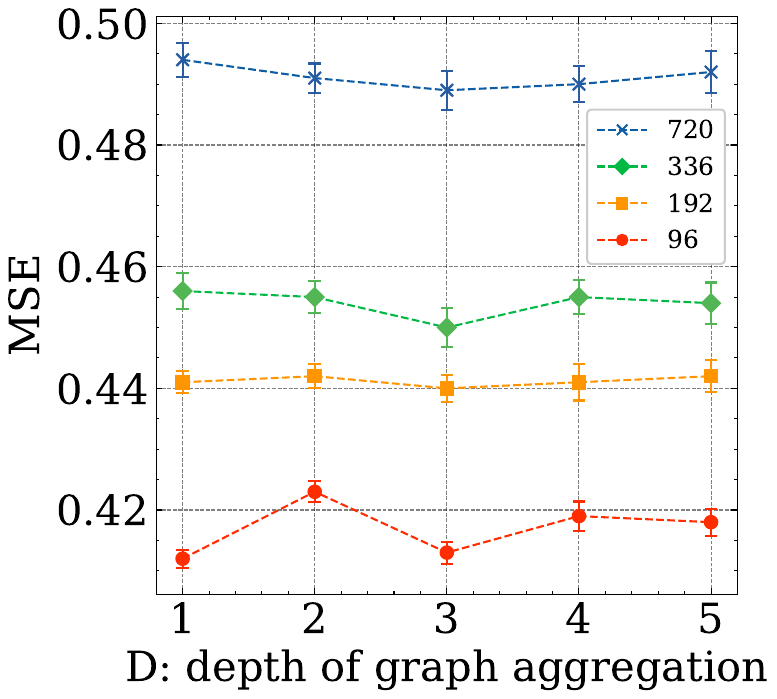}}%
\hfill
\subfloat[]{\label{fig:hyper_knn} \includegraphics[width=0.24\textwidth]{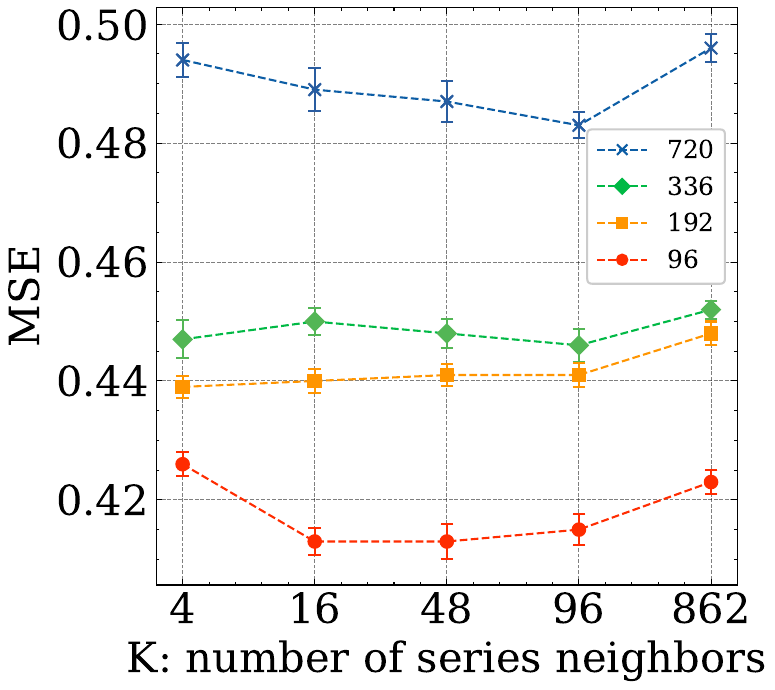}}%
\hfill
\subfloat[]{\label{fig:hyper_encoder_layers} \includegraphics[width=0.24\textwidth]{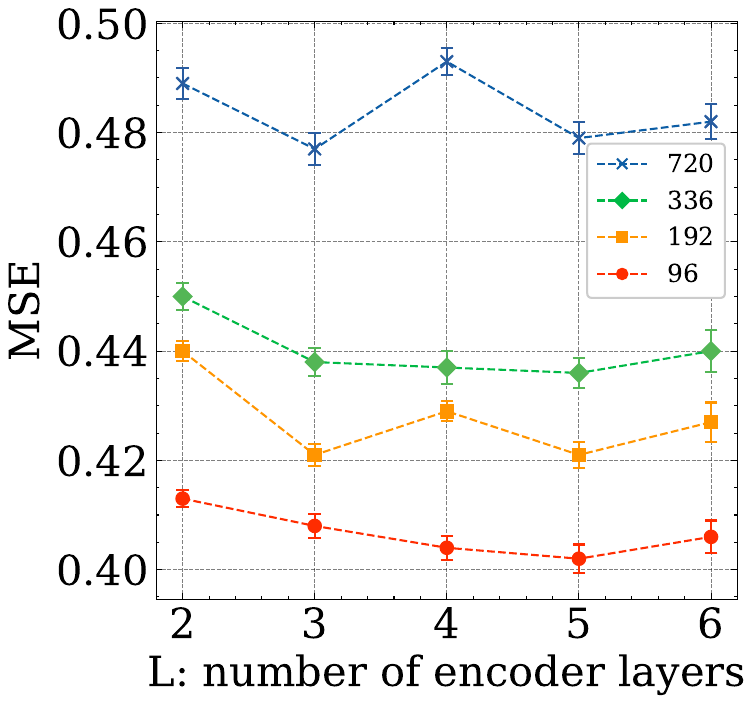}}%
\caption{Evaluation on hyper-parameter impact. (a) MSE against the length of global tokens on the Traffic dataset. (b) MSE against the graph aggregation depth on the Traffic dataset. (c) MSE against the number of nearest neighbors on the Traffic dataset. (d) MSE against SageFormer encoder layers on the Traffic dataset.}
\label{fig:hyper}
\end{figure*}

\subsection{Ablation Study}
The ablation studies were conducted to address two primary concerns: 1) the impact of graph aggregation and 2) the impact of series-aware global tokens. We designate SageFormer variants without specific components as shown in Table~\ref{tab:ablation_on_architecture}.

First, the experiments validated the effectiveness of the graph structure in our MTS forecasting model. Removing the graph aggregation module from each encoder layer resulted in a substantial decline in prediction accuracy. On the Traffic dataset, the average decrease was 7.3\%, and on the seven-series ETTh1 dataset, it was 2.8\%, showing that graph structures enhance performance more in datasets with numerous series. Second, series-aware global tokens enhanced the model's prediction accuracy while also reducing computational overhead. If all tokens (not just global tokens) participated in graph propagation calculations, the model's performance would decline by 6.3\% and 1.6\% on the Traffic and ETTh1 datasets, respectively. Lastly, we discovered that techniques like sparse constraints and directed graphs in graph construction were more effective for larger datasets (e.g., Traffic). In comparison, they had little impact on smaller datasets' prediction results (e.g., ETTs). This finding suggests that applying sparse constraints can mitigate the impact of variable redundancy on the model while conserving computational resources.

\subsection{Effect of Hyper-parameters}
In this section, we examine the impact of four hyperparameters on our proposed SageFormer model: global token length, depth of graph aggregation, the number of nearest neighbors, and the depth of encoder layers. We conduct a sensitivity analysis on the Traffic dataset (Fig.~\ref{fig:hyper}). For each of the four tasks, SageFormer consistently delivers stable performance, regardless of the selected value.

\hlcyan{\textbf{Global token length} (Fig.~\ref{fig:hyper_global_token}): The model's performance remains consistent across all prediction windows, irrespective of the value of $M$. To optimize computational efficiency, we set $M=1$.
\textbf{Depth of graph aggregation} (Fig.~\ref{fig:hyper_graph_depth}): The model demonstrates robust performance with varying graph aggregation depths. To balance accuracy and efficiency, we set $D=3$.
\textbf{Number of nearest neighbors} (Fig.~\ref{fig:hyper_knn}): Larger $K$ values generally yield better results, but performance declines when a fully connected graph is utilized. This suggests sequence redundancy in MTS forecasting tasks, so we select $K=16$.
\textbf{SageFormer encoder layers} (Fig.~\ref{fig:hyper_encoder_layers}): Increasing the number of encoding layers results in a higher parameter count for the model and its computational time. No significant reduction is observed after the model surpasses three layers, leading us to set the model's layers to $L=3$.
}

\hlcyan{Revisiting Global Token Length: Our study challenges the assumption that more global tokens always enhance model performance. Contrary to expectations, we found that increasing global token length does not linearly correlate with improved outcomes. This can be attributed to the efficacy of a single global token in capturing necessary intra-series interactions. Essentially, one global token is often adequate for encapsulating key series information, aligning with the principle of Occam's Razor: simpler solutions are preferable when effective. This insight led us to optimize the model with a single global token, balancing efficiency and performance.}

\begin{figure}[t]
\centering
\subfloat[]{\label{fig:synthetic_cycle} \includegraphics[width=0.485\columnwidth]{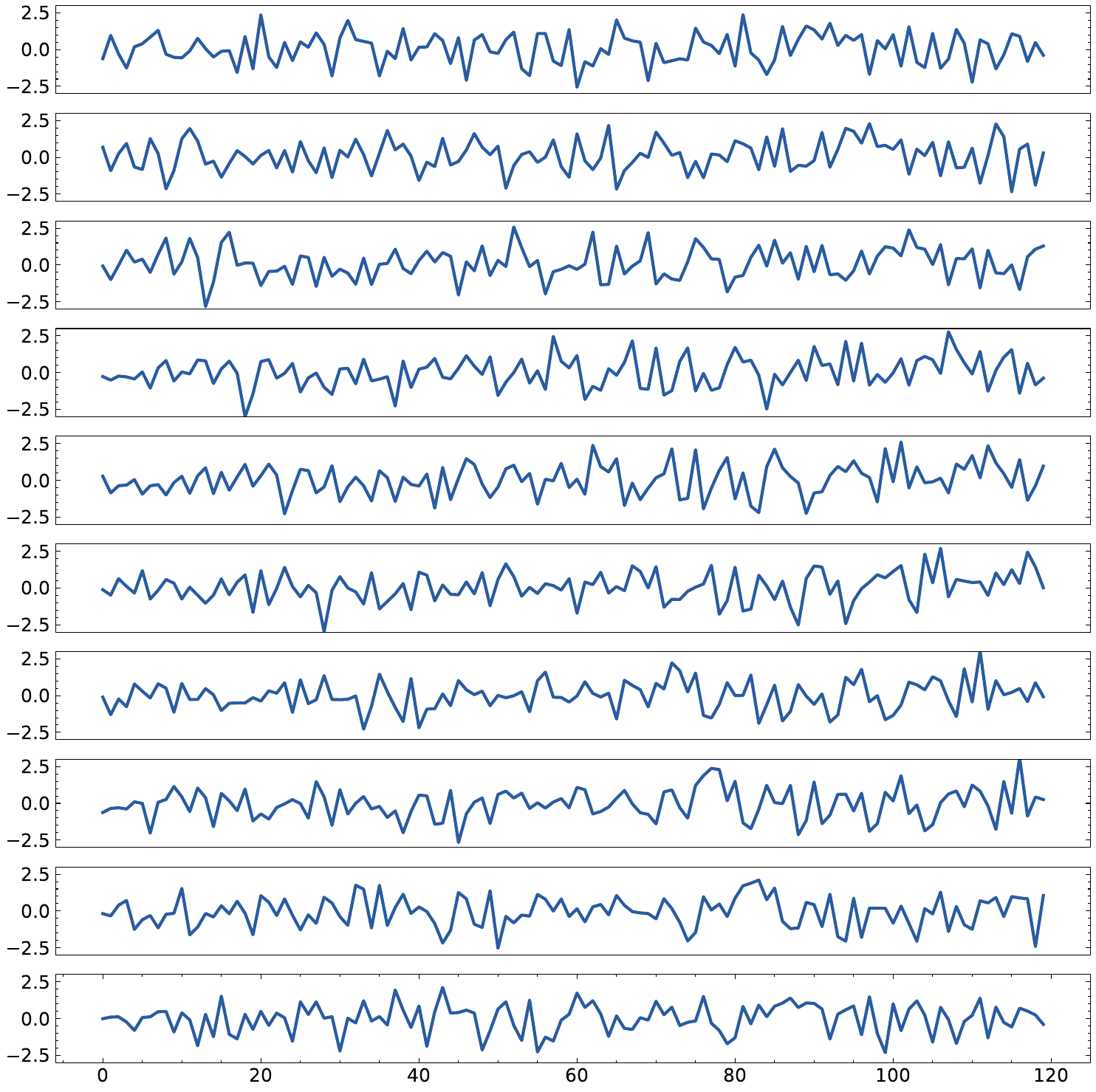}}
\hfill
\subfloat[]{\label{fig:synthetic_sin} \includegraphics[width=0.485\columnwidth]{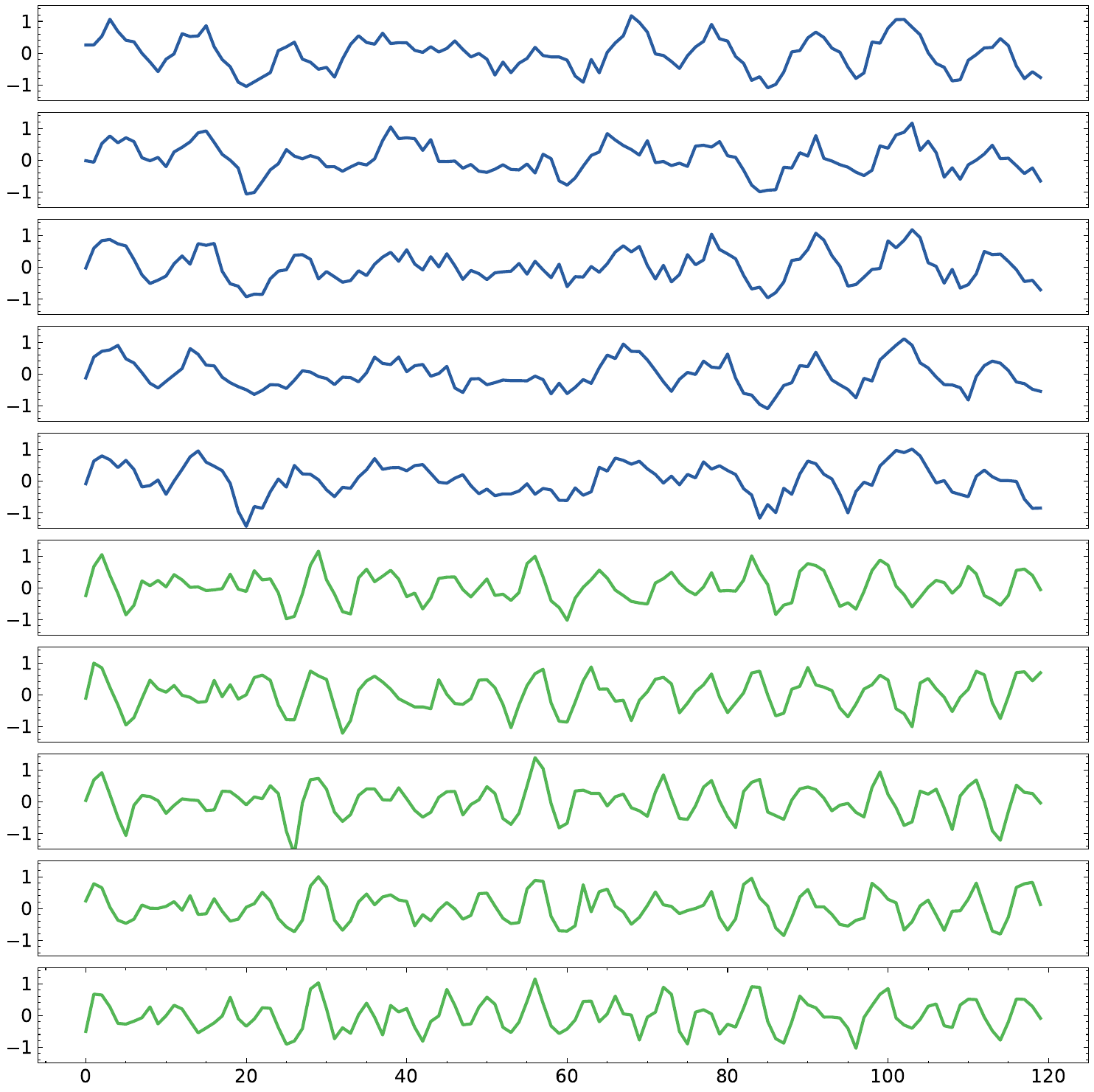}}%
\caption{Visualization of 120 timesteps of (a) the Directed Cycle Graph Dataset and (b) the Low-rank Dataset.}
\label{fig:synthetic_vis}
\end{figure}

\subsection{Synthetic Datasets} \label{sec:synthetic}

The design of synthetic datasets aims to emulate specific scenarios often encountered in real-world MTS data. Through two specialized datasets, we seek to benchmark SageFormer's performance and highlight the inherent challenges of different time series forecasting frameworks.

\begin{figure*}[t]
\centering
\subfloat[]{\label{fig:inferred_graph} \includegraphics[width=0.485\textwidth]{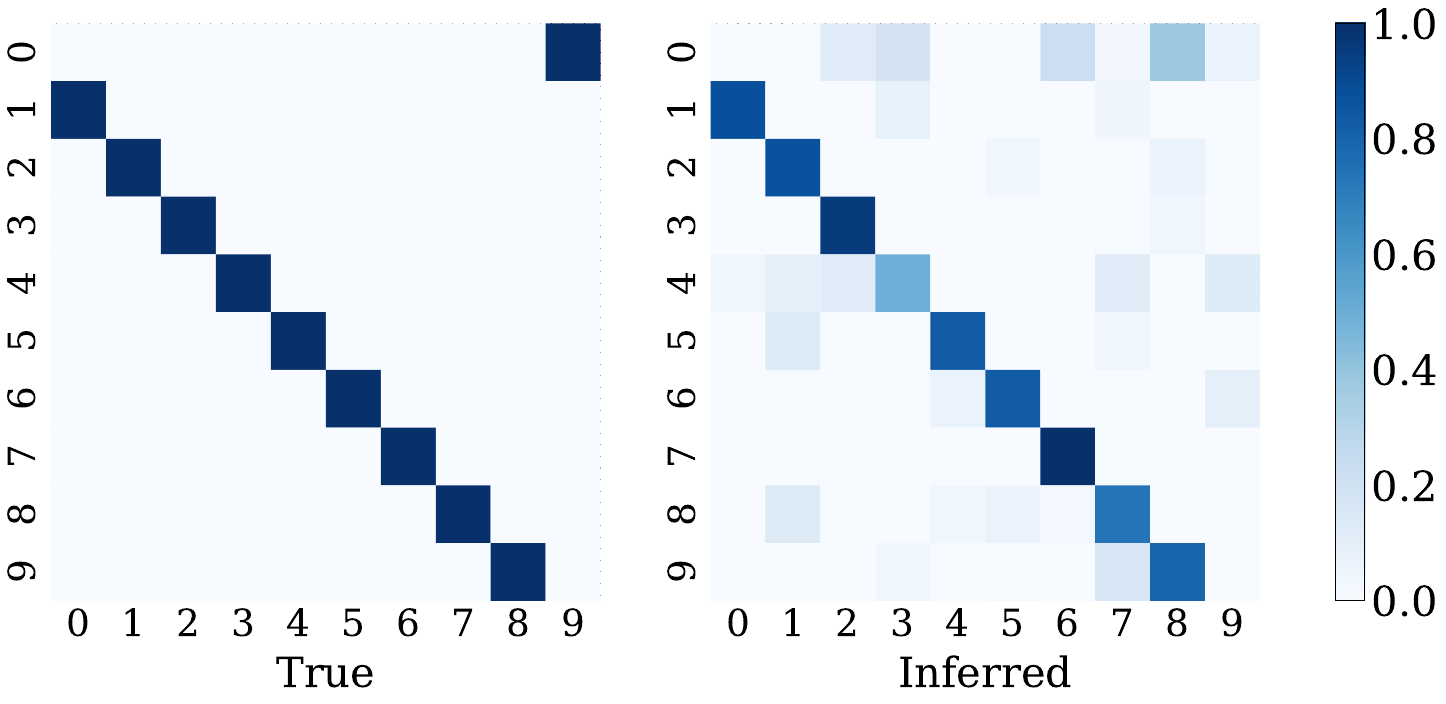}}
\hfill
\subfloat[]{\label{fig:cycle_mae} \includegraphics[width=0.238\textwidth]{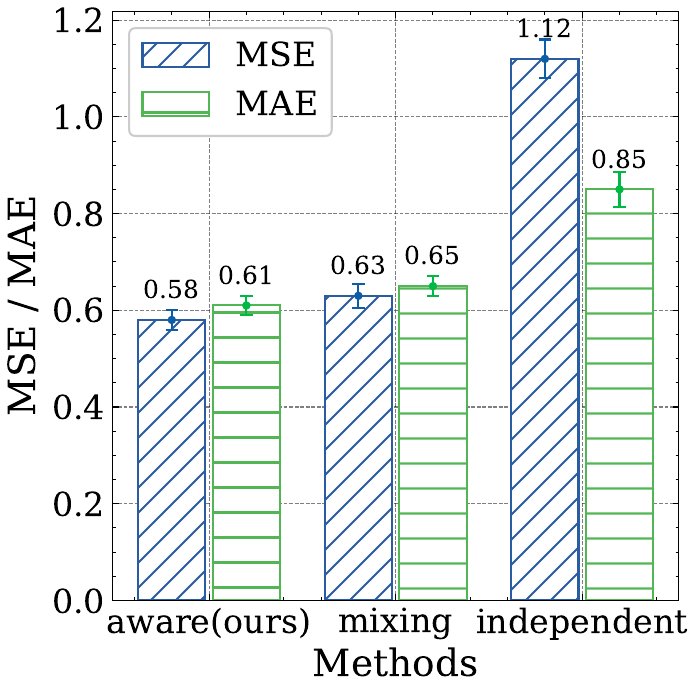}}%
\hfill
\subfloat[]{\label{fig:low-rank} \includegraphics[width=0.25\textwidth]{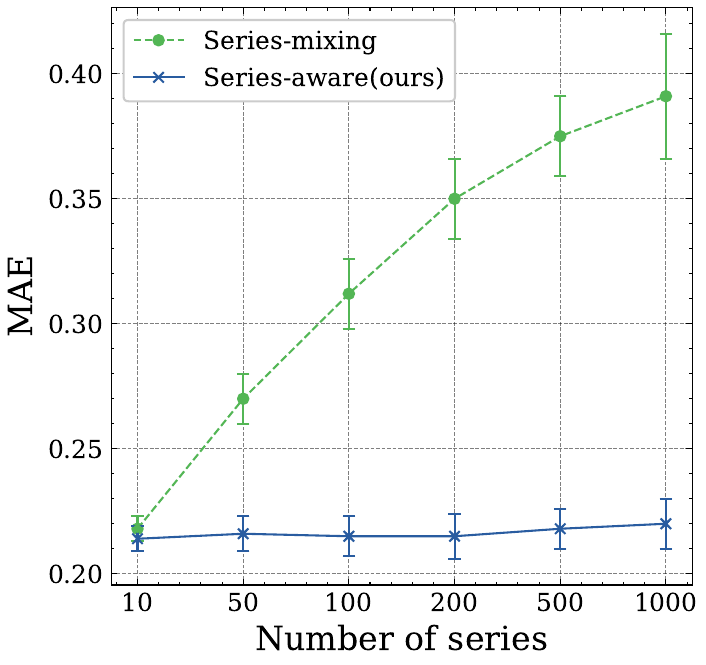}}%
\caption{Evaluation on synthetic datasets. (a) The left side displays the heat map of the actual adjacency matrix, while the right side presents the inferred adjacency matrix by SageFormer, illustrating the effectiveness of our proposed method in learning the inherent graph structure.; (b) Prediction results of three different methods on the Directed Cycle Graph dataset; \hlcyan{(c) Prediction MAE results for low-rank datasets with varying numbers of series ($N$)}. We selected the Nonstationary Transformer for the series-mixing method, and for the series-independent method, we chose PatchTST as a representative.}
\label{fig:synthetic}
\end{figure*}

\paragraph{Directed Cycle Graph Dataset} \label{sec:cycleGraph}

In this section, we assess the ability of SageFormer to infer adjacency matrices using a synthetic dataset characterized by a directed cycle graph structure (see Fig.~\ref{fig:synthetic_cycle}). The dataset contains a panel of \(N=10\) time series, each of 10,000 length. Uniquely, the value of each time series \(x_{i,t}\) is derived from the series indexed as \(i-1 \operatorname{mod} N\) with a temporal lag of \(\tau=10\).

Mathematically, the generation process is:

\begin{equation}
    x_{i, t} \sim \mathcal{N} \left(\beta x_{(i-1 \operatorname{mod} N), t-10}; \sigma^2\right),
\end{equation}
where \(x_{i, t}\) denotes the value of the \(i\)-th series at time \(t\). The sampling process uses a normal distribution, denoted by \(\mathcal{N}\), characterized by specific mean and variance parameters. The mean, represented as \(\beta x_{(i-1 \operatorname{mod} N), t-10}\), captures the scaled value from the previous series \(i-1 \operatorname{mod} N\) at time \(t-10\). Here, \(\beta\) serves as a scaling factor, and \(\sigma^2\) delineates the allowable variance.

This cyclical generation creates a directed cycle graph in the multivariate time series adjacency matrix, as illustrated in Fig.~\ref{fig:synthetic_cycle} for the initial 120 timesteps.
Fig.~\ref{fig:inferred_graph} juxtaposes the actual inter-series relationships with the inferences from our model, underscoring SageFormer's adeptness at recovering these connections. Our series-aware framework, as shown in Fig.~\ref{fig:cycle_mae}, demonstrates superior performance against previous series-mixing and series-independent models, registering the most minimal MAE and MSE test losses. Notably, the series-independent approach falters significantly in this context, recording an MSE above 1. This inadequacy can be attributed to its oversight of the prominent inter-series dependencies present in this dataset. Notably, within this dataset, each sub-series mirrors white noise characteristics.

\begin{figure}[t]
\centering
\includegraphics[width=\columnwidth]{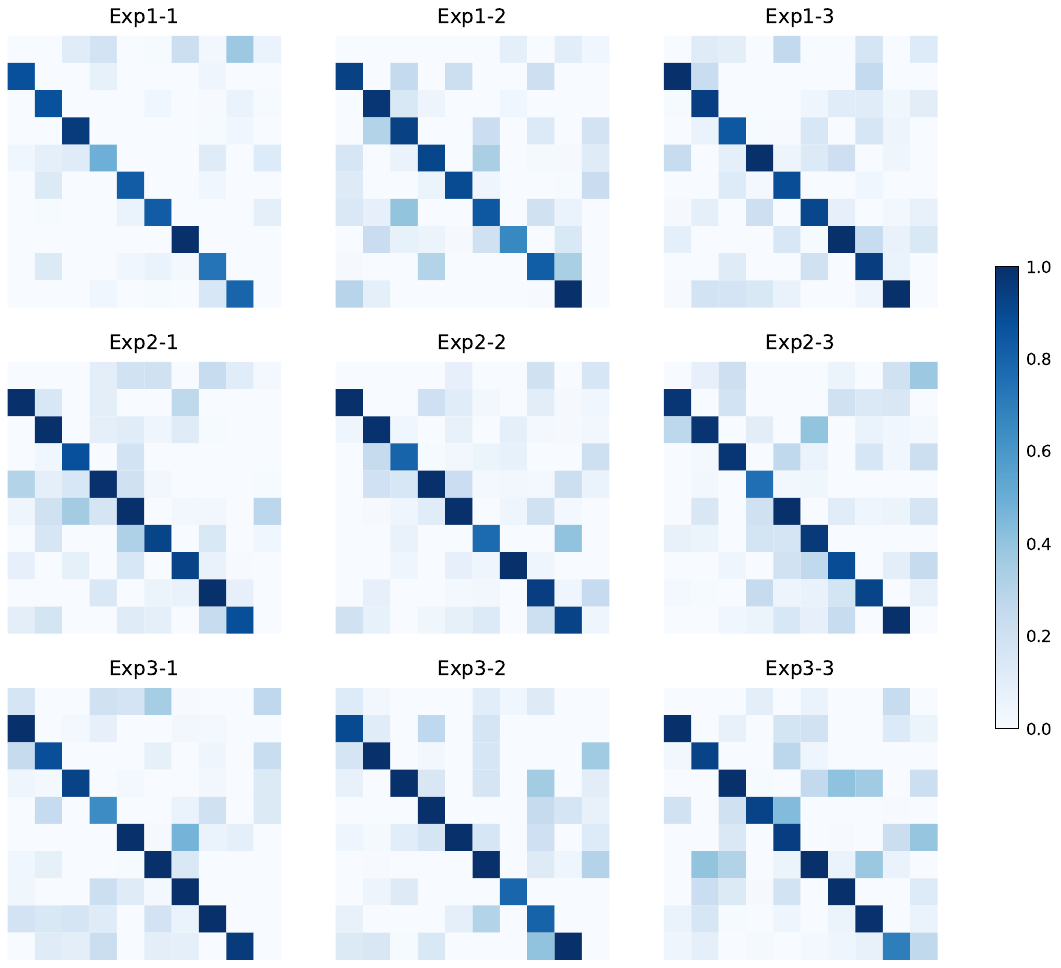}
\caption{\hlcyan{Graph structures inferred by the model from multiple random experiments. Each row represents an instance of a randomly constructed dataset, with the three images in the row showing the results from different random experiments, demonstrating the stability of the graph structure.}}
\label{fig:graph_stable}
\end{figure}

\begin{figure*}[t!]
\centering
\subfloat[series 1]{\label{fig:ETTh1_series1} \includegraphics[width=0.32\textwidth]{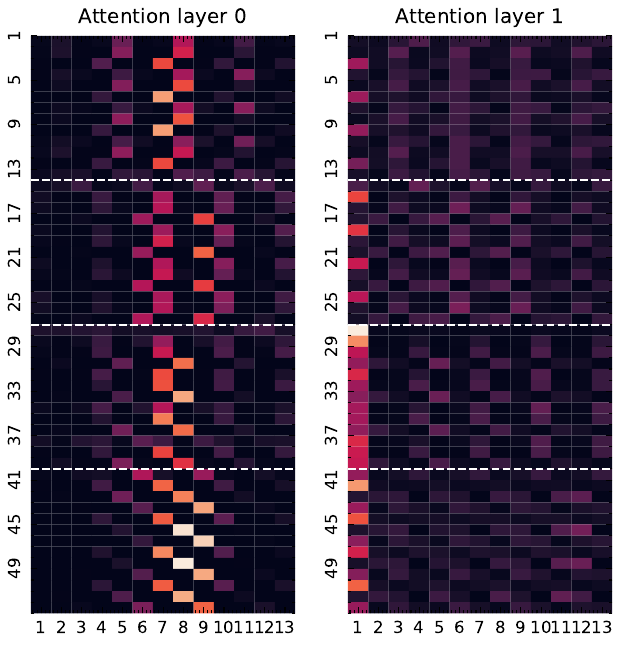}}
\hfill
\subfloat[series 3]{\label{fig:ETTh1_series3} \includegraphics[width=0.32\textwidth]{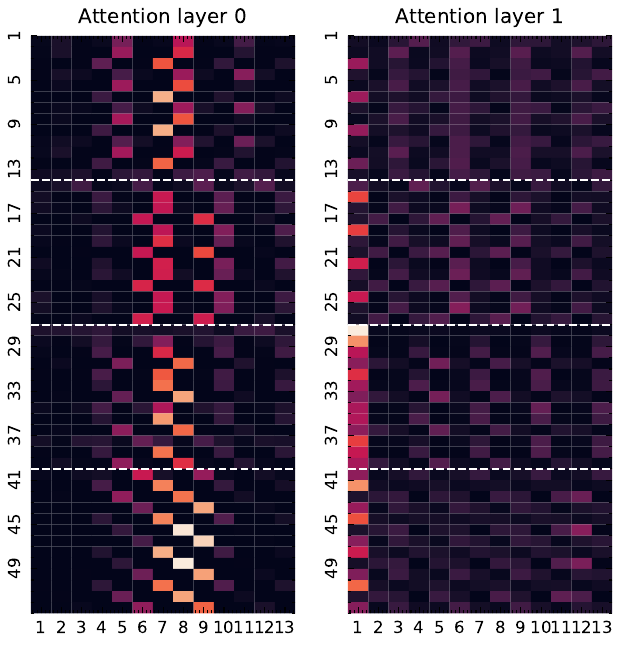}}%
\hfill
\subfloat[series 5]{\label{fig:ETTh1_series5} \includegraphics[width=0.32\textwidth]
{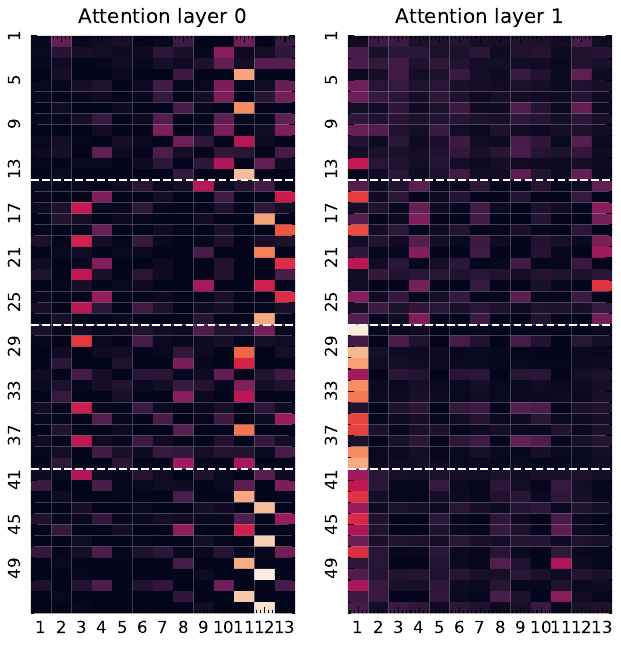}}%
\caption{\hlcyan{Multi-head attention weights on ETTh1 dataset. The weight dimensions for each layer are $4\times13, 13$, reflecting the multi-head attention mechanism applied to 12 tokens post patching and embedding, with the first token being the additionally introduced global token. Brighter indicates higher values.}}
\label{fig:ETT_attn}
\end{figure*}

\paragraph{Low-rank Dataset} \label{lowRank}
To investigate how various frameworks handle sparse data, we created several Low-rank MTS datasets with varying numbers of series (see Fig.~\ref{fig:synthetic_sin}). Drawing inspiration from the Discrete Sine Transformation, our datasets consist of signals synthesized from distinct sinusoids perturbed with Gaussian noise. The same sinusoids are shared among different groups of series, imbuing the datasets with the low-rank quality.

The synthetic dataset is constructed from time series, each of 10,000 length, and a variable number of series. Each series originates from a weighted combination of distinct sinusoids with specific frequencies and amplitudes. This creation process is encapsulated by:

\begin{equation}
    x_i = \sum_{m=1}^{M} B_{i,m} sin \left(2\pi\omega_{\lfloor i/K\rfloor,m}t\right)+\epsilon_{i,t},
\end{equation}
where $B_{i,m}$ is drawn from a normalized uniform distribution based on $\tilde{B}_{i,m} \sim U(0.4, 1)$. The frequency parameter, $\omega_{\lfloor i/K\rfloor,m}$, is sampled from $U(0, 0.2)$, while $\epsilon_{i,t}$ originates from $\mathcal{N}(0, 0.2^2)$. Crucially, $\omega_{\lfloor i/K\rfloor,m}$ remains consistent across series within a group, instilling the dataset's low-rank attribute. For our experiments, the chosen parameters were $M=3$ and $K=2$.

The dataset's distinctive low-rank feature is a consequence of the shared frequency values, $\omega_{\lfloor i/K\rfloor,m}$, among the series in a given group. This shared attribute allows the dataset to be captured with a succinct, low-dimensional representation, highlighting its low-rank nature.

Fig.~\ref{fig:low-rank} contrasts the prediction MAE of our method with the series-mixing approach for datasets of varying series numbers ($N$). It can be observed that the prediction performance of the series-mixing method deteriorates rapidly as the number of series increases since it encodes all series information into the same token. In contrast, the MAE of our method does not increase with the growth in the number of series, indicating that our designed approach can effectively exploit the low-rank characteristics of the dataset.

\subsection{Validity and Stability of the Graph}
\hlcyan{In this section, we aim to address a pivotal question: Are the graph structures derived from our proposed method valid and stable? The validity of the inferred graph structure is crucial for precise interpretation and prediction in Multivariate Time Series (MTS) data. Additionally, the stability of this structure is essential to ensure consistent model performance across varied datasets and conditions. However, a significant challenge arises due to the nature of the time series datasets used in our study. These datasets lack an actual, obtainable graph structure, and the dependency relationships in real data are often too complex for straightforward visualization and analysis. Consequently, we turn to experiments on synthetic datasets to conduct our analysis. }

\subsubsection{Validity of the Graph Structure}

\hlcyan{As previously mentioned in Section~\ref{sec:cycleGraph}, the validity of the graph structure has been tested using a Directed Cycle Graph Dataset. This synthetic dataset, designed to mirror real-world scenarios in MTS data, has been instrumental in demonstrating the model's capability to accurately infer the underlying graph structure. The adjacency matrices, both actual and inferred (as shown in Fig.~\ref{fig:inferred_graph}), highlight the effectiveness of the SageFormer in uncovering the inherent connections within the data.}

\subsubsection{Stability of the Graph Structure}

\hlcyan{In addition to the validation of the graph structure, we have undertaken extensive experiments to evaluate its stability. This involved generating visualizations for three different datasets across three independent trials, as depicted in Fig.~\ref{fig:graph_stable}. The striking similarity among the nine resultant graphs across these trials speaks volumes about the stability of the learned graph structure. This consistent reconstruction of true data dependency relationships, irrespective of the sample or trial, showcases the reliability of our model in various scenarios.

Our findings confirm that the graph structure inferred by our model is not only valid but also remarkably stable. The directed cycle graph dataset provides a clear validation of the graph structure, while the additional stability experiments reinforce the model's consistency and reliability. This dual assurance of validity and stability underlines the robustness of our approach and its effectiveness in leveraging complex inter-series relationships for MTS forecasting.}

\begin{figure*}[ht]
    \centering
    \includegraphics[width=\textwidth]{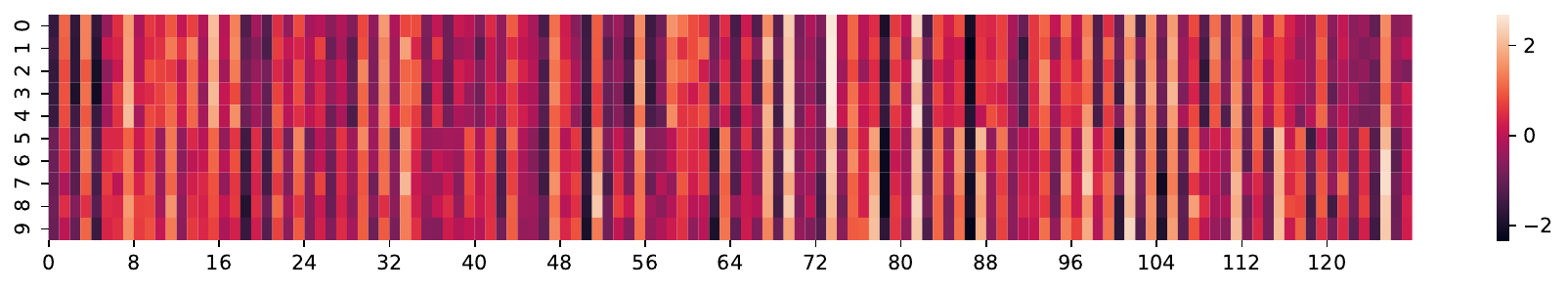}
    \caption{\hlcyan{Heatmap visualization of global tokens. The horizontal axis represents a 128-dimensional feature vector (sliced from the entire 512 dimensions), and the vertical axis represents the global tokens corresponding to the ten time series in the low-rank dataset.}}
    \label{fig:global_tokens}
\end{figure*}

\begin{figure*}[ht]
\centering
\subfloat[10 series]{\label{fig:tSNE_10} \includegraphics[width=0.3\textwidth]{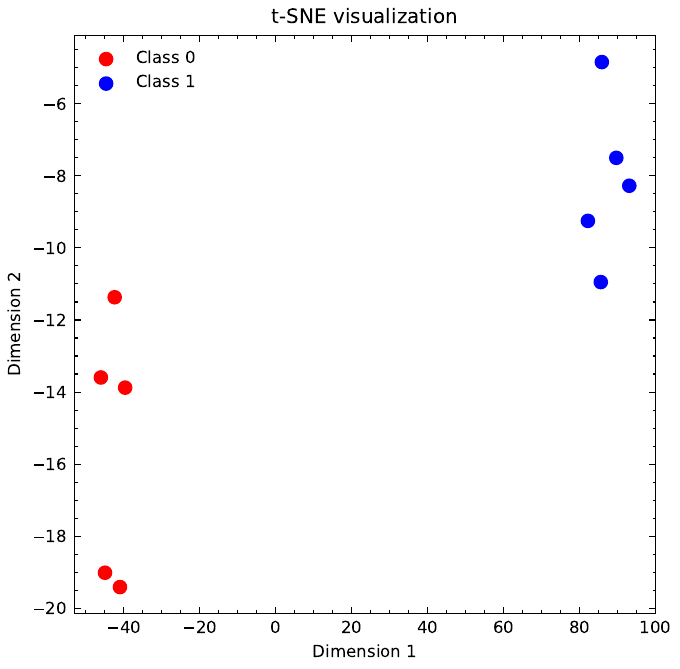}}
\hfill
\subfloat[50 series]{\label{fig:tSNE_50} \includegraphics[width=0.3\textwidth]{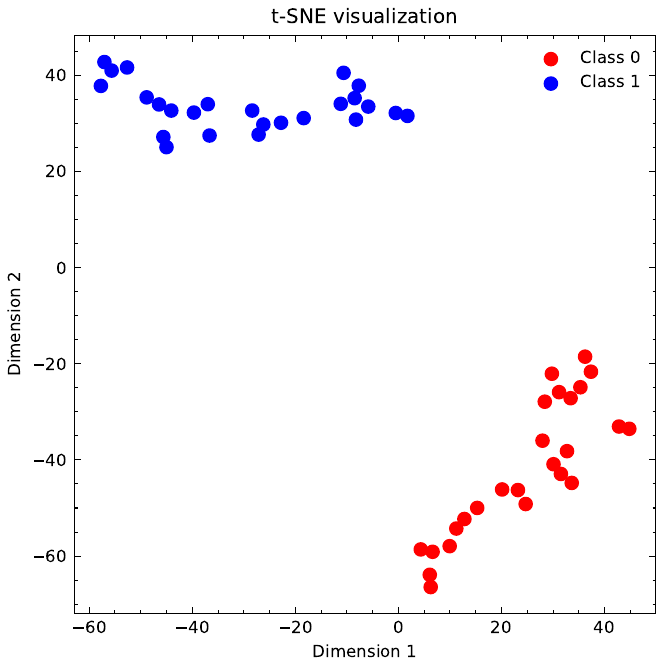}}%
\hfill
\subfloat[100 series]{\label{fig:tSNE_100} \includegraphics[width=0.3\textwidth]{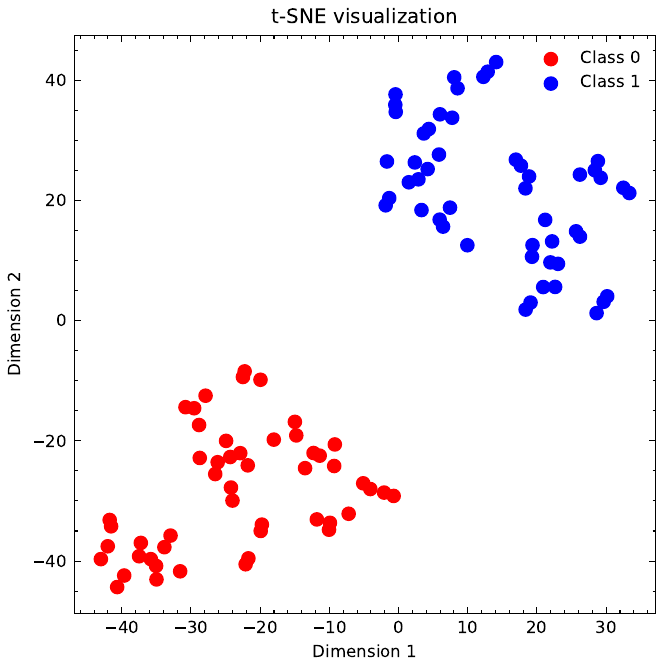}}%

\caption{\hlcyan{t-SNE visualization of global tokens in time series data of different dimensions.}}
\label{fig:tSNE}
\end{figure*}

\subsection{Discussion on Global Tokens} \label{sec:dis_on_global_tokens}

\subsubsection{Transformer Attention Weights Visualization}
\label{sec:c4exp1}
\hlcyan{To ascertain the effectiveness of global tokens in facilitating information transmission across distinct series, the Transformer attention weights were visualized for the ETTh1 dataset model. As illustrated in Figure~\ref{fig:ETT_attn}, each subfigure comprises a pair of heatmaps reflecting the attention weights in two different layers of the Transformer encoder.

We focus on the first column of attention weights, representing the temporal tokens' attention to the global token. Initially, attention towards the global token is low in the first layer due to its random initialization and replication across series. However, after the first layer, the global tokens start capturing intra-series dependencies, further learning inter-series dependencies through GNN propagation. Notably, in the second layer, the attention weights for the global token significantly increase, indicating its effective role in transmitting information across temporal tokens within each series. This contrast in attention weights between the layers confirms the global token's evolving functionality and the efficacy of our design.}

\subsubsection{Global Tokens Visualization}

\hlcyan{We expanded our analysis by visualizing global tokens on the synthesized low-rank datasets (see section~\ref{sec:synthetic} for the detailed construction process), selected for its complex inter-variable dynamics. Figure~\ref{fig:global_tokens}, showing a 10-variable dataset over 128 dimensions, presents a pronounced bifurcation into two groups, reflecting marked intra-group cohesion and inter-group distinctions. This bifurcation illustrates that global tokens, through GNN and transformer encoding, indeed effectively capture the variables' global characteristics.

To solidify our findings, we employed t-SNE for visualizing global tokens on three varied low-rank datasets with 10, 50, and 100 variables. Fig.~\ref{fig:tSNE} consistently demonstrates clear segregation of the two groups, confirming the global tokens' capability, coupled with the graph structure, to adeptly capture both intra- and inter-series dependencies.

In conclusion, these additional experiments and visualizations provide a clear, intuitive understanding of our model's ability to effectively discern and represent complex inter-series dependencies and redundancies, underscoring the integrated global tokens and learned graph structure's effectiveness.}

\subsection{Computational Efficiency Analysis}
To further investigate the training overhead compared to other baselines, we compared the computational efficiency of our model, SageFormer, with other Transformer-based models (Table~\ref{tab:complexity}). Although SageFormer's complexity is theoretically quadratic to historical series length $T$, a large patch length $P$ in practice brings its runtime close to linear complexity models. An additional $O(C^2)$ complexity is due to standard graph convolution operations, but techniques exist to reduce this to linear complexity~\cite{wu2019simplifying, levie2018cayleynets}. In the decoder part, the complexity of SageFormer is simplified to linear, owing to the streamlined design of the linear decoder head.

We also evaluated running time and memory consumption on the Traffic dataset, which has the most variables. SageFormer balances running time and memory usage well, achieving a running time of $0.31\pm0.03$ seconds per batch and consuming $12.42$ GB of memory. This result is slightly slower compared to the PatchTST~\cite{nie2022time} model but is faster than the Crossformer~\cite{zhang2023crossformer} model. These results suggest that our proposed SageFormer model presents a competitive trade-off between efficiency and prediction accuracy. 

\begin{table*}[hbt]
    \caption{Computational Complexity Per Layer of Transformer-based Models.}\label{tab:complexity}
     \centering
     \begin{threeparttable}
    \begin{tabular}{c|cccc} \toprule
    Methods & Encoder layer & Decoder layer & Time(s/batch) & Memory(GB) \\ 
    \hline
    Transformer \cite{NIPS2017_3f5ee243} & $O(T^2)$ & $O\left(\tau(\tau+T)\right)$ & $0.07\pm0.01$ & $2.39$\Tstrut \\
    Informer \cite{haoyietal-informer-2021} & $O(T\log T)$ & $O\left(\tau(\tau+\log T)\right)$ & $0.04\pm0.04$ & $2.31$ \\
    FEDformer \cite{zhou2022fedformer} & $O(T)$ & $O\left(\tau+T/2\right)$ & $0.18\pm0.03$ & $3.13$ \\
    PatchTST \cite{nie2022time} & $O(CT^2/P^2)$ & $O(\tau)$ & $0.22\pm0.04$ & $11.44$ \\
    Crossformer \cite{zhang2023crossformer} & $O(CT^2/P^2)$ & $O(C\tau(\tau+T)/P^2)$ & $0.82\pm0.05$ & $22.53$\Bstrut\\ 
    \hline 
    SageFormer \textbf{(ours)} & $O(CT^2/P^2+C^2)$ & $O(\tau)$ & $0.31\pm0.03$ & $12.42$ \Tstrut \\
    \bottomrule
    \end{tabular}

    \begin{tablenotes}
        \item[] $T$ denotes the length of the historical series, $\tau$ represents the length of the prediction window, $C$ is the number of series, and $P$ corresponds to the segment length of each patch.
    \end{tablenotes}
    \end{threeparttable}
\end{table*}

\begin{table*}[th]
	\centering
	\caption{\hlcyan{Performance Promotion Full Results.}}\label{tab:add-on-transformer}
   \begin{threeparttable}
    \begin{tabular}{c|c|cc|cc|cc|cc|cc|cc} 
    \toprule
    \multicolumn{2}{c}{Dataset} & \multicolumn{2}{c}{Transformer} & \multicolumn{2}{c}{\textbf{+ Series-aware} \tnote{2}} & \multicolumn{2}{c}{Informer \tnote{3}} & \multicolumn{2}{c}{\textbf{+ Series-aware}} & \multicolumn{2}{c}{FEDformer} & \multicolumn{2}{c}{\textbf{+ Series-aware}}\\  
    \cmidrule(lr){3-4} \cmidrule(lr){5-6} \cmidrule(lr){7-8} \cmidrule(lr){9-10} \cmidrule(lr){11-12} \cmidrule(lr){13-14}
    \multicolumn{2}{c}{Metric} & MSE & MAE & MSE & MAE & MSE & MAE & MSE & MAE & MSE & MAE & MSE & MAE\\ 
    \toprule
    \multirow{5}{*}{\rotatebox{90}{\scalebox{0.95}{Traffic}}}
    &  96 & 0.644 & 0.354 & \boldres{0.526} & \boldres{0.281} & 0.587 & 0.366 & \boldres{0.561} & \boldres{0.318} & 0.719 & 0.391 & \boldres{0.672} & \boldres{0.382} \\ 
    & 192 & 0.662 & 0.365 & \boldres{0.536} & \boldres{0.287} & 0.604 & 0.373 & \boldres{0.572} & \boldres{0.320} & 0.696 & 0.379 & \boldres{0.688} & \boldres{0.369} \\ 
    & 336 & 0.661 & 0.361 & \boldres{0.553} & \boldres{0.296} & 0.621 & 0.383 & \boldres{0.582} & \boldres{0.321} & 0.777 & 0.420 & \boldres{0.708} & \boldres{0.405} \\ 
    & 720 & 0.677 & 0.370 & \boldres{0.581} & \boldres{0.306} & 0.626 & 0.382 & \boldres{0.598} & \boldres{0.312} & 0.864 & 0.472 & \boldres{0.761} & \boldres{0.428} \\ 
    \cmidrule(lr){2-14}
    & Avg & 0.661 & 0.362 & \boldres{0.549} & \boldres{0.293} & 0.610 & 0.376 & \boldres{0.578} & \boldres{0.318} & 0.764 & 0.416 & \boldres{0.707} & \boldres{0.396} \\ 
    \midrule

    \multirow{5}{*}{\rotatebox{90}{\scalebox{0.95}{Electricity}}}
    &  96 & 0.259 & 0.358 & \boldres{0.186} & \boldres{0.272} & 0.193 & 0.308 & \boldres{0.182} & \boldres{0.278} & 0.274 & 0.368 & \boldres{0.188} & \boldres{0.275} \\ 
    & 192 & 0.265 & 0.363 & \boldres{0.188} & \boldres{0.275} & 0.201 & 0.315 & \boldres{0.191} & \boldres{0.285} & 0.296 & 0.386 & \boldres{0.212} & \boldres{0.311} \\ 
    & 336 & 0.273 & 0.369 & \boldres{0.203} & \boldres{0.293} & 0.214 & 0.329 & \boldres{0.206} & \boldres{0.313} & 0.300 & 0.394 & \boldres{0.223} & \boldres{0.327} \\ 
    & 720 & 0.291 & 0.377 & \boldres{0.234} & \boldres{0.320} & 0.246 & 0.355 & \boldres{0.248} & \boldres{0.345} & 0.373 & 0.439 & \boldres{0.267} & \boldres{0.361} \\ 
    \cmidrule(lr){2-14}
    & Avg & 0.272 & 0.367 & \boldres{0.202} & \boldres{0.290} & 0.214 & 0.327 & \boldres{0.207} & \boldres{0.305} & 0.311 & 0.397 & \boldres{0.223} & \boldres{0.319} \\ 
    \midrule
      
    \multirow{5}{*}{\rotatebox{90}{\scalebox{0.95}{Weather}}}
    &  96 & 0.399 & 0.424 & \boldres{0.189} & \boldres{0.252} & 0.217 & 0.296 & \boldres{0.208} & \boldres{0.276} & 0.300 & 0.384 & \boldres{0.178} & \boldres{0.224} \\ 
    & 192 & 0.566 & 0.537 & \boldres{0.235} & \boldres{0.299} & 0.276 & 0.336 & \boldres{0.249} & \boldres{0.312} & 0.598 & 0.544 & \boldres{0.226} & \boldres{0.262} \\ 
    & 336 & 0.631 & 0.582 & \boldres{0.295} & \boldres{0.359} & 0.339 & 0.380 & \boldres{0.296} & \boldres{0.322} & 0.578 & 0.523 & \boldres{0.287} & \boldres{0.305} \\ 
    & 720 & 0.849 & 0.685 & \boldres{0.440} & \boldres{0.481} & 0.403 & 0.428 & \boldres{0.385} & \boldres{0.382} & 1.059 & 0.741 & \boldres{0.384} & \boldres{0.423} \\ 
    \cmidrule(lr){2-14}
    & Avg & 0.611 & 0.557 & \boldres{0.290} & \boldres{0.348} & 0.309 & 0.360 & \boldres{0.285} & \boldres{0.323} & 0.634 & 0.548 & \boldres{0.269} & \boldres{0.304} \\ 
    \midrule

    \multirow{5}{*}{\rotatebox{90}{\scalebox{0.95}{ETTh1}}}
    &  96 & 0.780 & 0.685 & \boldres{0.384} & \boldres{0.403} & 0.376 & 0.419 & \boldres{0.377} & \boldres{0.408} & 0.865 & 0.713 & \boldres{0.578} & \boldres{0.512} \\ 
    & 192 & 0.906 & 0.755 & \boldres{0.435} & \boldres{0.433} & 0.420 & 0.448 & \boldres{0.431} & \boldres{0.440} & 1.008 & 0.792 & \boldres{0.669} & \boldres{0.558} \\ 
    & 336 & 0.980 & 0.797 & \boldres{0.479} & \boldres{0.459} & 0.459 & 0.465 & \boldres{0.455} & \boldres{0.456} & 1.107 & 0.809 & \boldres{0.699} & \boldres{0.592} \\ 
    & 720 & 1.008 & 0.800 & \boldres{0.538} & \boldres{0.528} & 0.506 & 0.507 & \boldres{0.467} & \boldres{0.462} & 1.181 & 0.865 & \boldres{0.745} & \boldres{0.644} \\ 
    \cmidrule(lr){2-14}
    & Avg & 0.919 & 0.759 & \boldres{0.459} & \boldres{0.456} & 0.440 & 0.460 & \boldres{0.433} & \boldres{0.442} & 1.040 & 0.795 & \boldres{0.673} & \boldres{0.577} \\ 
    \bottomrule
	\end{tabular}

    \begin{tablenotes}
        \item[1] We compare base models and the proposed series-aware framework under different prediction lengths (96, 192, 336, 720). The input sequence length is set to 512 for all datasets. \textbf{Bold} indicates the \textbf{better}.
        \item[2] This model diverges from SageFormer in that only the latter employs patching to generate temporal tokens.
        \item[3] The inconsistency between the Informer results in Table \ref{tab:full_forecasting_results} and Table \ref{tab:add-on-transformer} arises because of different input history lengths.
    \end{tablenotes}
  \end{threeparttable}
\end{table*}

\begin{table*}[htbp]
	\caption{\hlcyan{Long-Term Forecasting Task Full Results.}}\label{tab:full_forecasting_results}
	\centering
	\begin{threeparttable}
	\renewcommand{\multirowsetup}{\centering}
    \renewcommand\TPTtagStyle{\textit}
	\setlength{\tabcolsep}{2.6pt}
	\begin{tabular}{c|c|cc|cc|cc|cc|cc|cc|cc|cc|cc|cc}
	  \toprule
	  \multicolumn{2}{c}{\multirow{2}{*}{Models}} & 
	  \multicolumn{2}{c}{\rotatebox{0}{\cellcolor{red!25}\textbf{SageFormer}}} & 
	  \multicolumn{2}{c}{\rotatebox{0}{\cellcolor{blue!25}\scalebox{0.92}{Crossformer\tnote{a}}}} & 
	  \multicolumn{2}{c}{\rotatebox{0}{\cellcolor{blue!25}MTGNN\tnote{a}}} & 
	  \multicolumn{2}{c}{\rotatebox{0}{\cellcolor{blue!25}LSTnet\tnote{a}}} & 
	  \multicolumn{2}{c}{\rotatebox{0}{\cellcolor{green!25}PatchTST\tnote{b}}} & 
	  \multicolumn{2}{c}{\rotatebox{0}{\cellcolor{green!25}DLinear\tnote{b}}} & 
	  \multicolumn{2}{c}{\rotatebox{0}{\cellcolor{yellow!25}Stationary\tnote{c}}} & 
	  \multicolumn{2}{c}{\rotatebox{0}{\cellcolor{yellow!25}Autoformer\tnote{c}}} & 
	  \multicolumn{2}{c}{\rotatebox{0}{\cellcolor{yellow!25}Informer\tnote{c}}} & 
	  \multicolumn{2}{c}{\rotatebox{0}{\cellcolor{yellow!25}Transformer\tnote{c}}}  
	  \\
	  \multicolumn{2}{c}{} & 
	  \multicolumn{2}{c}{\cellcolor{red!25}(\textbf{Ours})} & 
	  \multicolumn{2}{c}{\cellcolor{blue!25}\cite{zhang2023crossformer}\tnote{*}} &
	  \multicolumn{2}{c}{\cellcolor{blue!25}\cite{wu2020connecting}\tnote{*}} &
	  \multicolumn{2}{c}{\cellcolor{blue!25}\cite{2018Modeling}\tnote{*}} &
	  \multicolumn{2}{c}{\cellcolor{green!25}\cite{nie2022time}\tnote{*}} &
	  \multicolumn{2}{c}{\cellcolor{green!25}\cite{Zeng2022AreTE}\tnote{*}} &
	  \multicolumn{2}{c}{\cellcolor{yellow!25}\cite{Liu2022NonstationaryTR}} &
	  \multicolumn{2}{c}{\cellcolor{yellow!25}\cite{wu2021autoformer}} & 
	  \multicolumn{2}{c}{\cellcolor{yellow!25}\cite{haoyietal-informer-2021}} &
	  \multicolumn{2}{c}{\cellcolor{yellow!25}\cite{NIPS2017_3f5ee243}} \\
  
	  \cmidrule(lr){3-4} \cmidrule(lr){5-6}\cmidrule(lr){7-8} \cmidrule(lr){9-10}\cmidrule(lr){11-12}\cmidrule(lr){13-14}\cmidrule(lr){15-16}\cmidrule(lr){17-18}\cmidrule(lr){19-20}\cmidrule(lr){21-22}
	  \multicolumn{2}{c}{Metric}&MSE&MAE&MSE&MAE&MSE&MAE&MSE&MAE&MSE&MAE&MSE&MAE&MSE&MAE&MSE&MAE&MSE&MAE&MSE&MAE\\
	  \toprule
	  \multirow{5}{*}{\rotatebox{90}{\scalebox{0.95}{Traffic}}} 
	  & 96&\boldres{0.408}&\boldres{0.271}&0.544&0.307&0.574&0.312&0.711&0.432&\secondres{0.450}&\secondres{0.287}&0.650&0.396&0.612&0.338&0.613&0.388&0.719&0.391&0.644&0.354\\
	  &192&\boldres{0.421}&\boldres{0.279}&0.566&0.315&0.587&0.315&0.722&0.441&\secondres{0.456}&\secondres{0.292}&0.598&0.370&0.613&0.340&0.616&0.382&0.696&0.379&0.662&0.365\\
	  &336&\boldres{0.438}&\boldres{0.283}&0.571&0.314&0.594&0.318&0.741&0.451&\secondres{0.471}&\secondres{0.297}&0.605&0.373&0.618&0.328&0.622&0.337&0.777&0.420&0.661&0.361\\
	  &720&\boldres{0.477}&\boldres{0.308}&0.599&0.313&0.612&0.322&0.768&0.474&\secondres{0.509}&\secondres{0.317}&0.645&0.394&0.653&0.355&0.660&0.408&0.864&0.472&0.677&0.370\\
	  \cmidrule(lr){2-22}
	  &Avg&\boldres{0.436}&\boldres{0.285}&0.570&0.312&0.592&0.317&0.736&0.450&\secondres{0.472}&\secondres{0.298}&0.625&0.383&0.624&0.340&0.628&0.379&0.764&0.416&0.661&0.363\\
	  \midrule
  
	  \multirow{5}{*}{\rotatebox{90}{\scalebox{0.95}{Electricity}}} 
	  & 96&\boldres{0.147}&\boldres{0.246}&0.213&0.300&0.243&0.342&0.382&0.452&0.175&\secondres{0.266}&0.197&0.282&\secondres{0.169}&0.273&0.201&0.317&0.274&0.368&0.259&0.358\\
	  &192&\boldres{0.161}&\boldres{0.259}&0.290&0.351&0.298&0.364&0.401&0.482&0.184&\secondres{0.274}&0.196&0.285&\secondres{0.182}&0.286&0.222&0.334&0.296&0.386&0.265&0.363\\
	  &336&\boldres{0.180}&\boldres{0.279}&0.348&0.389&0.368&0.396&0.419&0.477&0.200&\secondres{0.290}&0.209&0.301&\secondres{0.200}&0.304&0.231&0.338&0.300&0.394&0.273&0.369\\
	  &720&\boldres{0.213}&\boldres{0.309}&0.405&0.425&0.422&0.410&0.556&0.565&0.240&0.322&0.245&0.333&\secondres{0.222}&\secondres{0.321}&0.254&0.361&0.373&0.439&0.291&0.377\\
	  \cmidrule(lr){2-22}
	  &Avg&\boldres{0.175}&\boldres{0.273}&0.314&0.366&0.333&0.378&0.440&0.494&0.200&\secondres{0.288}&0.212&0.300&\secondres{0.193}&0.296&0.227&0.338&0.311&0.397&0.272&0.367\\
	  \midrule
  
	  \multirow{5}{*}{\rotatebox{90}{\scalebox{0.95}{Weather}}}
	  & 96&\boldres{0.162}&\boldres{0.206}&\boldres{0.162}&0.232&0.189&0.252&0.682&0.594&0.175&\secondres{0.216}&0.196&0.255&\secondres{0.173}&0.223&0.266&0.336&0.300&0.384&0.399&0.424\\
	  &192&\secondres{0.211}&\boldres{0.250}&\boldres{0.208}&0.277&0.235&0.299&0.755&0.652&\secondres{0.219}&\secondres{0.256}&0.237&0.296&0.245&0.285&0.307&0.367&0.598&0.544&0.566&0.537\\
	  &336&\secondres{0.271}&\boldres{0.294}&\boldres{0.265}&0.320&0.295&0.359&0.782&0.683&0.277&\secondres{0.297}&0.283&0.335&0.321&0.338&0.359&0.395&0.578&0.523&0.631&0.582\\
	  &720&\boldres{0.345}&\boldres{0.343}&0.388&0.391&0.440&0.481&0.851&0.757&\secondres{0.353}&\secondres{0.346}&\boldres{0.345}&0.381&0.414&0.410&0.419&0.428&1.059&0.741&0.849&0.685\\
	  \cmidrule(lr){2-22}
	  &Avg&\boldres{0.247}&\boldres{0.273}&\secondres{0.256}&0.305&0.290&0.348&0.768&0.672&\secondres{0.256}&\secondres{0.279}&0.265&0.317&0.288&0.314&0.338&0.382&0.634&0.548&0.611&0.557\\
	  \midrule
  
	  \multirow{5}{*}{\rotatebox{90}{\scalebox{0.95}{ETTm1}}}
	  & 96&\secondres{0.324}&\secondres{0.362}&0.361&0.402&0.428&0.446&1.339&0.913&\boldres{0.319}&\boldres{0.361}&0.345&0.372&0.386&0.398&0.505&0.475&0.672&0.571&0.701&0.609\\
	  &192&\boldres{0.368}&\boldres{0.387}&0.403&0.440&0.457&0.469&1.542&1.009&\secondres{0.370}&\secondres{0.389}&0.380&0.389&0.459&0.444&0.553&0.496&0.795&0.669&0.829&0.676\\
	  &336&\boldres{0.401}&\boldres{0.408}&0.551&0.535&0.579&0.562&1.920&1.234&\secondres{0.406}&\secondres{0.410}&0.413&0.413&0.495&0.464&0.621&0.537&1.212&0.871&1.024&0.783\\
	  &720&\boldres{0.457}&\boldres{0.441}&0.720&0.649&0.798&0.671&2.987&1.669&\secondres{0.461}&\secondres{0.444}&0.474&0.453&0.585&0.516&0.671&0.561&1.166&0.823&1.189&0.843\\
	  \cmidrule(lr){2-22}
	  &Avg&\boldres{0.388}&\boldres{0.400}&0.509&0.507&0.566&0.537&1.947&1.206&\secondres{0.389}&\secondres{0.401}&0.403&0.407&0.481&0.456&0.588&0.517&0.961&0.734&0.936&0.728\\
	  \midrule
  
	  \multirow{5}{*}{\rotatebox{90}{\scalebox{0.95}{ETTm2}}}
	  & 96&\boldres{0.173}&\boldres{0.255}&0.253&0.347&0.289&0.364&0.723&0.655&\secondres{0.177}&\secondres{0.261}&0.193&0.292&0.192&0.274&0.255&0.339&0.365&0.453&0.515&0.534\\
	  &192&\boldres{0.239}&\boldres{0.299}&0.421&0.483&0.456&0.492&1.285&0.932&\secondres{0.241}&\secondres{0.303}&0.284&0.362&0.280&0.339&0.281&0.340&0.533&0.563&1.424&0.892\\
	  &336&\boldres{0.299}&\boldres{0.338}&1.276&0.805&1.432&0.812&3.064&1.556&\secondres{0.302}&\secondres{0.343}&0.369&0.427&0.334&0.361&0.339&0.372&1.363&0.887&1.183&0.829\\
	  &720&\boldres{0.395}&\boldres{0.395}&3.783&1.354&2.972&1.336&5.484&1.978&\secondres{0.401}&\secondres{0.403}&0.554&0.522&0.417&0.413&0.433&0.432&3.379&1.338&2.788&1.237\\
	  \cmidrule(lr){2-22}
	  &Avg&\boldres{0.277}&\boldres{0.322}&1.433&0.747&1.287&0.751&2.639&1.280&\secondres{0.280}&\secondres{0.328}&0.350&0.401&0.306&0.347&0.327&0.371&1.410&0.810&1.478&0.873\\
	  \midrule
  
	  \multirow{5}{*}{\rotatebox{90}{\scalebox{0.95}{ETTh1}}}
	  & 96&\secondres{0.377}&\secondres{0.397}&0.421&0.448&0.522&0.490&1.654&0.982&\boldres{0.372}&\boldres{0.396}&0.386&0.400&0.513&0.491&0.449&0.459&0.865&0.713&0.780&0.685\\
	  &192&\boldres{0.423}&\boldres{0.425}&0.534&0.515&0.542&0.536&1.999&1.218&\secondres{0.439}&\secondres{0.433}&0.437&0.432&0.534&0.504&0.500&0.482&1.008&0.792&0.906&0.755\\
	  &336&\boldres{0.459}&\boldres{0.445}&0.656&0.581&0.736&0.643&2.655&1.369&\secondres{0.468}&\secondres{0.456}&0.481&0.459&0.588&0.535&0.521&0.496&1.107&0.809&0.980&0.797\\
	  &720&\boldres{0.465}&\boldres{0.466}&0.849&0.709&0.916&0.750&2.143&1.380&\secondres{0.491}&\secondres{0.486}&0.519&0.516&0.643&0.616&0.514&0.512&1.181&0.865&1.008&0.800\\
	  \cmidrule(lr){2-22}
	  &Avg&\boldres{0.431}&\boldres{0.433}&0.615&0.563&0.679&0.605&2.113&1.237&\secondres{0.443}&\secondres{0.443}&0.456&0.452&0.570&0.537&0.496&0.487&1.040&0.795&0.919&0.759\\
	  \midrule
  
	  \multirow{5}{*}{\rotatebox{90}{\scalebox{0.95}{ETTh2}}}
	  & 96&\boldres{0.286}&\boldres{0.338}&1.142&0.772&1.843&0.911&3.252&1.604&\secondres{0.302}&\secondres{0.346}&0.333&0.387&0.476&0.458&0.346&0.388&3.755&1.525&2.590&1.282\\
	  &192&\boldres{0.368}&\boldres{0.394}&1.784&1.022&2.439&1.325&5.122&2.309&\secondres{0.379}&\secondres{0.398}&0.477&0.476&0.512&0.493&0.456&0.452&5.602&1.931&6.464&2.102\\
	  &336&\boldres{0.413}&\boldres{0.429}&2.644&1.404&2.944&1.405&4.051&1.742&\secondres{0.418}&\secondres{0.431}&0.594&0.541&0.552&0.551&0.482&0.486&4.721&1.835&5.851&1.970\\
	  &720&\secondres{0.427}&\secondres{0.449}&3.111&1.501&3.244&1.592&5.104&2.378&\boldres{0.423}&\boldres{0.440}&0.831&0.657&0.562&0.560&0.515&0.511&3.647&1.625&3.063&1.411\\
	  \cmidrule(lr){2-22}
	  &Avg&\boldres{0.374}&\boldres{0.403}&2.170&1.175&2.618&1.308&4.382&2.008&\secondres{0.381}&\secondres{0.404}&0.559&0.515&0.526&0.516&0.450&0.459&4.431&1.729&4.492&1.691\\
	  \midrule
  
	  \multirow{5}{*}{\rotatebox{90}{\scalebox{0.95}{Exchange}}} 
	  & 96&\boldres{0.082}&\boldres{0.200}&0.256&0.367&0.289&0.388&0.593&0.711&\secondres{0.087}&\secondres{0.205}&0.088&0.218&0.111&0.237&0.197&0.323&0.847&0.752&0.527&0.566\\
	  &192&\boldres{0.175}&\boldres{0.298}&0.469&0.508&0.441&0.498&1.164&1.047&0.183&\secondres{0.305}&\secondres{0.176}&0.315&0.219&0.335&0.300&0.369&1.204&0.895&0.942&0.737\\
	  &336&0.331&\secondres{0.416}&0.901&0.741&0.934&0.774&1.544&1.196&\secondres{0.321}&\boldres{0.410}&\boldres{0.313}&0.427&0.421&0.476&0.509&0.524&1.672&1.036&1.485&0.946\\
	  &720&\secondres{0.826}&\secondres{0.683}&1.398&0.965&1.478&1.037&3.425&1.833&\boldres{0.825}&\boldres{0.680}&0.839&0.695&1.092&0.769&1.447&0.941&2.478&1.310&2.588&1.341\\
	  \cmidrule(lr){2-22}
	  &Avg&\boldres{0.354}&\boldres{0.399}&0.756&0.645&0.786&0.674&1.681&1.197&\boldres{0.354}&\secondres{0.400}&\boldres{0.354}&0.414&0.461&0.454&0.613&0.539&1.550&0.998&1.386&0.898\\
	  \midrule
  
	  \multirow{5}{*}{\rotatebox{90}{\scalebox{0.95}{ILI}}}
	  & 24&\secondres{2.180}&\secondres{0.858}&3.110&1.179&4.265&1.387&4.975&1.660&\boldres{2.066}&\boldres{0.854}&2.398&1.040&2.294&0.945&3.483&1.287&5.764&1.677&4.644&1.434\\
	  & 36&2.214&\boldres{0.877}&3.429&1.222&4.777&1.496&5.322&1.659&\secondres{2.173}&\secondres{0.881}&2.646&1.088&\boldres{1.825}&0.848&3.103&1.148&4.755&1.467&4.533&1.442\\
	  & 48&2.068&\boldres{0.876}&3.451&1.203&5.333&1.592&5.425&1.632&\secondres{2.032}&\secondres{0.892}&2.614&1.086&\boldres{2.010}&0.900&2.669&1.085&4.763&1.469&4.758&1.469\\
	  & 60&\secondres{1.991}&\boldres{0.898}&3.678&1.255&5.070&1.552&5.477&1.675&\boldres{1.987}&\secondres{0.899}&2.804&1.146&2.178&0.963&2.770&1.125&5.264&1.564&5.199&1.540\\
	  \cmidrule(lr){2-22}
	  &Avg&2.113&\boldres{0.877}&3.417&1.215&4.861&1.507&5.300&1.657&\boldres{2.065}&\secondres{0.882}&2.616&1.090&\secondres{2.077}&0.914&3.006&1.161&5.137&1.544&4.784&1.471\\
	  \midrule
	  \multicolumn{2}{c}{{$1^{\text{st}}$ Count}}&\multicolumn{2}{c}{55}&\multicolumn{2}{c}{3}&\multicolumn{2}{c}{0}&\multicolumn{2}{c}{0}&\multicolumn{2}{c}{12}&\multicolumn{2}{c}{2}&\multicolumn{2}{c}{2}&\multicolumn{2}{c}{0}&\multicolumn{2}{c}{0}&\multicolumn{2}{c}{0}\\
	  \bottomrule
	\end{tabular}
	  \begin{tablenotes}
          \item[1] We compare extensive competitive models under different prediction lengths (96, 192, 336, 720). The input sequence length is set to 36 for the ILI dataset and 96 for the others. \emph{Avg} is averaged from all four prediction lengths. \textbf{Bold}/\underline{underline} indicates the \textbf{best}/\underline{second}. 
          \item [2] \textit{a} marks the models explicitly utilizing inter-series dependencies; \textit{b} marks series-independent neural models; \textit{c} marks series-mixing transformer-based models. 
		  \item[3] $\ast$ means that there are some mismatches between our input-output setting and their papers. We adopt their official codes and only change the length of input and output sequences for a fair comparison.
	\end{tablenotes}

\end{threeparttable}
\end{table*}

\section{Conclusion}
This paper presented the series-aware framework and SageFormer, a novel approach for modeling both intra- and inter-series dependencies in long-term MTS forecasting tasks. By amalgamating GNNs with Transformer structures, SageFormer can effectively capture diverse temporal patterns and harness dependencies among various series. Our model has demonstrated impressive versatility through extensive experimentation, delivering state-of-the-art performance on real-world and synthetic datasets. SageFormer thus presents a promising solution to overcome the limitations of series dependencies modeling in MTS forecasting tasks and exhibits potential for further advancements and applications in other domains involving inter-series dependencies.

We also acknowledged the limitations of our work and briefly delineated potential avenues for future research. While SageFormer achieves exceptional performance in long-term MTS forecasting, the dependencies it captures do not strictly represent causality. As a result, some dependencies may prove unreliable in practical scenarios due to the non-stationary nature of time series. Our primary focus on enhancing long-term forecasting performance has led to some degree of overlooking the interpretability of the graph structure. Moving forward, our work's graph neural network component could be improved to learn causal relationships between variables and reduce its complexity. The framework proposed in this paper could also be applied to non-Transformer models in the future.


%



\appendix

\hlcyan{In the appendix, we present the full experimental results that complement the discussions and analyses provided in the main body of the paper. The detailed results offer a comprehensive view of the performance of our proposed SageFormer model compared to other baselines across various datasets and settings.
Table~\ref{tab:full_forecasting_results} provides the complete set of results for long-term forecasting, extending the summary presented in Table~\ref{tab:long_term_forecasting_results_short}. And Table~\ref{tab:add-on-transformer} extends the discussion from the main text (see Table~\ref{tab:add-on-transformer_short}) by providing a complete set of results when applying the SageFormer framework to various Transformer-based architectures.}

\section*{Acknowledgment}

The authors would like to thank Xin Wang and Leon Yan for their technical supports. The authors also extend their gratitude to all the reviewers for their valuable comments and suggestions that greatly improved the manuscript. The usual disclaimer applies.

\ifCLASSOPTIONcaptionsoff
  \newpage
\fi

\newpage


\bibliographystyle{IEEEtran}
\bibliography{IEEEabrv,references}
\end{document}